\pdfoutput=1

\documentclass[11pt]{article}

\usepackage[]{acl} %

\usepackage[english]{babel}

\usepackage{graphicx}

\usepackage{times}
\usepackage{latexsym}

\usepackage[T1]{fontenc}

\usepackage[utf8]{inputenc}

\usepackage{microtype}

\usepackage{inconsolata}

\usepackage{adjustbox}
\usepackage{booktabs}
\usepackage{multirow}
\usepackage{amsmath}
\usepackage{amssymb}
\usepackage{bm}
\usepackage{soul}
\usepackage{tcolorbox}
\usepackage[frozencache,cachedir=minted-cache]{minted}
\usepackage{subcaption}
\usepackage[justification=justified,skip=5pt]{caption}
\usepackage{comment}
\usepackage{makecell}

\usepackage{twemojis}
\usepackage{animate}
\usepackage{times}
\usepackage{tikz}
\usetikzlibrary{arrows.meta}
\usetikzlibrary{backgrounds}
\usetikzlibrary{positioning, fit, mindmap, trees, calc,tikzmark,shapes}
\usetikzlibrary{shapes.arrows, fadings, automata,tikzmark,decorations.pathreplacing,patterns}
\usetikzlibrary{decorations.text}

\usepackage{pgfplots}
\usepgfplotslibrary{groupplots}
\pgfplotsset{compat=1.14}
\usepackage{tikzpeople}

\usepackage{graphicx}

\usepackage{sistyle}
\usepackage{paralist}
\usepackage[inline]{enumitem}
\usepackage{amsmath}
\SIthousandsep{,}
\usepackage[normalem]{ulem} %

\usepackage{url}

\usepackage{xcolor}

\definecolor{mine_red}{RGB}{239, 129, 131}
\definecolor{mine_blue}{RGB}{105, 158, 212}
\definecolor{mine_green}{RGB}{164, 217, 187}
\definecolor{mine_font}{RGB}{0, 128, 0}
\definecolor{light_mine_blue}{RGB}{145, 198, 252}
\definecolor{light_mine_red}{RGB}{255, 149, 151}

\definecolor{riptide}{RGB}{141,211,199}
\definecolor{pale_prim}{RGB}{255,255,179}
\definecolor{lavender_gray}{RGB}{190,186,218}
\definecolor{salmon}{RGB}{242,131,107}
\definecolor{seagull}{RGB}{128,177,211}
\definecolor{rajah}{RGB}{253,180,98}
\definecolor{yellow_green}{RGB}{198,222,119}
\definecolor{classic_rose}{RGB}{252,205,229}
\definecolor{feijoa}{RGB}{178,223,138}

\definecolor{cruise}{RGB}{179,226,205}
\definecolor{apricot}{RGB}{253,205,172}
\definecolor{periwinkle}{RGB}{203,213,232}
\definecolor{snow_flurry}{RGB}{230,245,201}
\definecolor{buttermilk}{RGB}{255,242,174}

\definecolor{sundown}{RGB}{249, 180, 181}
\definecolor{spindle}{RGB}{179,205,227}
\definecolor{tea_green}{RGB}{204,235,197}
\definecolor{languid_lavender}{RGB}{222,203,228}
\definecolor{champagne}{RGB}{254,217,166}
\definecolor{cream}{RGB}{255,255,204}

\definecolor{monte_carlo}{RGB}{135,204,194}
\definecolor{melon}{RGB}{254,191,181}
\definecolor{granny_smith_apple}{RGB}{150,214,150}
\definecolor{watusi}{RGB}{254,221,207}
\definecolor{see_green}{RGB}{161,228,195}

\definecolor{moss_green}{RGB}{170,216,176}
\definecolor{opal}{RGB}{164,207,190}

\definecolor{pale_turquoise}{RGB}{172,240,242}
\definecolor{Madang}{RGB}{190,235,159}
\definecolor{pixie_green}{RGB}{183,214,170}
\definecolor{coral_andy}{RGB}{243,204,205}
\definecolor{manhattan}{RGB}{226,180,125}
\definecolor{quartz}{RGB}{219,223,238}
\definecolor{spring_sun}{RGB}{242,243,195}
\definecolor{dairy_cream}{RGB}{254,226,189}
\definecolor{surf_crest}{RGB}{205,230,208}
\definecolor{french_pass}{RGB}{195,232,246}
\definecolor{cosmos}{RGB}{248,209,210}
\definecolor{portafino}{RGB}{245,237,160}
\definecolor{sail}{RGB}{163,205,235}
\definecolor{hint_green}{RGB}{226,246,209}

\definecolor{jet_stream}{RGB}{188, 214, 210}

\definecolor{azalea}{RGB}{251, 196, 196}
\definecolor{wewak}{RGB}{244, 143, 150}
\definecolor{bittersweet}{RGB}{255,111,105}
\definecolor{sunset_orange}{RGB}{242,89,75}
\definecolor{light_coral}{RGB}{244, 127, 123}
\definecolor{carnation}{RGB}{245, 80, 86}
\definecolor{flamingo}{RGB}{237, 88, 85}
\definecolor{carmine_pink}{RGB}{231, 76, 60}
\definecolor{deep_carmine_pink}{RGB}{236, 50, 67}
\definecolor{fire_engine_red}{RGB}{210,44,41}
\definecolor{amaranth}{RGB}{234,46,73}
\definecolor{ku_crimson}{RGB}{243, 0, 25}
\definecolor{fire_engine_red}{RGB}{206, 37, 51}
\definecolor{copper_rust}{RGB}{155, 64, 74}

\definecolor{chilean_fire}{RGB}{215, 87, 44}

\definecolor{japanese_laurel}{RGB}{53, 116, 40}

\definecolor{turmeric}{RGB}{211, 178, 76}
\definecolor{saffron}{RGB}{249,193,62}
\definecolor{my_sin}{RGB}{255, 176, 59}
\definecolor{tree_poppy}{RGB}{246, 154, 27}
\definecolor{jaffa}{RGB}{240, 131, 58}
\definecolor{crusta}{RGB}{254, 127, 44}
\definecolor{tahiti_gold}{RGB}{223, 102, 36}
\definecolor{outrageous_orange}{RGB}{255, 100, 45}
\definecolor{safety_orange}{RGB}{254, 106, 0}

\definecolor{turquoise}{RGB}{41,217,194}
\definecolor{puerto_rico}{RGB}{94, 194, 166}
\definecolor{mountain_meadow}{RGB}{0, 163, 136}
\definecolor{free_speech_aquamarine}{RGB}{0, 156, 114}
\definecolor{java}{RGB}{2,190,196}

\definecolor{matisse}{RGB}{25, 104, 167}
\definecolor{shakespeare}{RGB}{85, 154, 193}
\definecolor{mona_lisa}{RGB}{246,152,134}

\definecolor{bgc}{RGB}{245,245,245}
\definecolor{tuatara}{RGB}{67, 67, 67}
\definecolor{aluminum}{RGB}{153,153,153}
\definecolor{silver}{RGB}{191,191,191}
\definecolor{platinum}{RGB}{228,228,228}
\definecolor{mercury}{RGB}{230,230,230}
\definecolor{gallery}{RGB}{240,240,240}
\definecolor{athens_gray}{RGB}{236, 240, 241}
\definecolor{ship_gray}{RGB}{77,77,77}

\definecolor{early_dawn}{RGB}{252,243,218}
\definecolor{egg_shell}{RGB}{238, 234, 215}
\definecolor{midnight}{RGB}{0, 29, 50}
\definecolor{sundown}{RGB}{249, 180, 181}
\definecolor{sun_shade}{RGB}{255, 144, 68}
\definecolor{sushi}{RGB}{117, 168, 47}
\definecolor{tomato}{RGB}{255, 97, 56}
\definecolor{ice_cold}{RGB}{169,232,220}

\definecolor{jelly_bean}{RGB}{45, 126, 150}
\definecolor{celestial_blue}{RGB}{52, 152, 219}
\definecolor{curious_blue}{RGB}{41, 128, 185}
\definecolor{french_blue}{RGB}{0, 112, 182}
\definecolor{matisse}{RGB}{25, 104, 167}

\definecolor{biscay}{RGB}{44, 62, 80}

\definecolor{cosmic_latte}{RGB}{222, 247, 229}
\definecolor{chinook}{RGB}{163, 232, 178}
\definecolor{padua}{RGB}{121, 189, 143}
\definecolor{ocean_green}{RGB}{79, 176, 112}
\definecolor{pastel_green}{RGB}{107, 227, 135}
\definecolor{chateau_green}{RGB}{69, 191, 85}
\definecolor{RoyalBlue}{RGB}{69, 191, 85}
\definecolor{pigment_green}{RGB}{0, 175, 79}
\definecolor{fern}{RGB}{101,197,117}
\definecolor{killarney}{RGB}{56, 113, 66}
\definecolor{viridian}{RGB}{70, 137, 102}

\newcommand{\halfwing}[1]{
    \begin{scope}[yscale=1,xscale=#1]
        \filldraw[fill=black!90!white!,draw=black,thick,rounded corners=1mm] (0,0.2) -- (0,1.4) -- (-2,1.4) .. controls (-4,0.8) .. (-4.3,0.2) .. controls (-4.48,0.08) .. (-4.5,-0.15) .. controls (-4.9,-0.5) and (-4.9,-0.7) .. (-4.7,-0.9) .. controls (-4.7,-1) .. (-4.6,-1.1) .. controls (-4.9,-1.8) .. (-4.2,-2) -- (-4,-2.4) .. controls (-4.1,-3) .. (-3.6,-3.1) -- (-3.25,-3.7) .. controls (-3.5,-4.5) .. (-4.1,-5.4) .. controls (-4.2,-5.9) and (-3.6,-5.9) .. (-3.5,-5.4) .. controls (-3.55,-5.1) and (-3.4,-4.8) .. (-3,-4.1) -- (-2.6,-4.1) .. controls (-2.35,-4.35) .. (-2,-4.2) .. controls (-1.75,-4.6) and (-1.25,-4.6) .. (-1.25,-3.90) .. controls (-0.9,-4) .. (-0.6,-2.8) -- (-0.3,-1) -- (0,0.2);
        \shadedraw[top color=blue!45!cyan!,bottom color=blue!20!cyan!,draw=black,rounded corners=1mm] (-0.8,-3) .. controls (-0.5,-2) .. (-0.30,-0.95){[rounded corners=0mm] .. controls (-0.15,-0.3) .. (-0.05,0.45) -- (-0.05,0.7)} -- (-0.7,0.4) .. controls (-0.9,-2) .. (-0.8,-3);
        \shadedraw[top color=blue!60!cyan!,bottom color=blue!20!cyan!,draw=black] (-1.1,-2.9) .. controls (-1.4,2) and (0.5,2) .. (-1.1,-2.9);
        \shadedraw[top color=blue!70!cyan!,bottom color=blue!20!cyan!,draw=black] (-1.6,-2.8) .. controls (-0.9,3.7) and (0,-0.1) .. (-1.6,-2.8);
        \shadedraw[top color=blue!70!cyan!,bottom color=blue!20!cyan!,draw=black] (-2.1,-2.6) .. controls (-0.9,3.7) and (-0.3,-0.1) .. (-2.1,-2.6);
        \shadedraw[top color=blue!80!cyan!,bottom color=blue!20!cyan!,draw=black] (-2.6,-2.2) .. controls (-0.3,3.7) and (-0.3,-0.1) .. (-2.6,-2.2);
        \shadedraw[top color=blue!80!cyan!,bottom color=blue!20!cyan!,draw=black] (-3,-1.7) .. controls (0.1,3.7) and (0.1,-0.1) .. (-3,-1.7);
        \shadedraw[top color=blue!80!cyan!,bottom color=blue!20!cyan!,draw=black] (-3.4,-1.2) .. controls (0.77,3) and (0.77,-0.2) .. (-3.4,-1.2);
        \shadedraw[top color=blue!80!cyan!,bottom color=blue!20!cyan!,draw=black] (-3.6,-0.6) .. controls (0.77,2.2) and (0.77,-0.5) .. (-3.6,-0.6);
        \shadedraw[top color=blue!80!cyan!,bottom color=blue!15!cyan!,draw=black] (-3.5,0) .. controls (0.77,1.8) and (0.77,-0.2) .. (-3.5,0);
        \shadedraw[top color=blue!80!cyan!,bottom color=blue!10!cyan!,draw=black] (-2.5,0.7) .. controls (0.77,2) and (0.77,0) .. (-2.5,0.7);
        \shadedraw[top color=blue!45!cyan!,bottom color=blue!15!cyan!,draw=black] (-0.05,0.6) -- (-0.05,0.9) .. controls (-4,-0.5) and (-1.5,-2) .. (-0.05,0.6);

        \filldraw[fill=black!90!white!,draw=black,thick] (0,1) -- (0,2.2) [rounded corners=7mm] parabola[bend at end] (-6,6) -- (-5,1) -- (0,1);
        \shadedraw[top color=blue!20!cyan!,bottom color=blue!70!cyan!,draw=black] (-4,3.5) .. controls (3,-1) and (-2,4) .. (-4,3.5);
        \shadedraw[top color=blue!15!cyan!,bottom color=blue,draw=black] (-4.1,3) .. controls (3.9,-0.5) and (-2.1,4) .. (-4.1,3);
        \shadedraw[top color=blue!15!cyan!,bottom color=blue!80!cyan!,draw=black] (-4.2,2.5) .. controls (4,-0.2) and (-2.2,3.5) .. (-4.2,2.5);
        \shadedraw[top color=blue!15!cyan!,bottom color=blue!70!cyan!,draw=black] (-4.2,1.8) .. controls (4,0.3) and (-2.2,3) .. (-4.2,1.8);
        \shadedraw[top color=cyan,bottom color=blue!60!cyan!,draw=black] (-4.2,1.2) .. controls (4.05,0.9) and (-2.2,2.4) .. (-4.2,1.2);
        \shadedraw[top color=cyan,bottom color=blue!60!cyan!,draw=black] (-0.05,1.85) -- (-0.05,1.80) .. controls (-6.5,6.5) and (-2,5.5) .. (-0.05,1.85);
        \shadedraw[top color=blue!10!cyan!,bottom color=blue!50!cyan!,draw=black] (-0.05,1.8) -- (-0.05,1.4) .. controls (-7.5,5.5) and (-2,5) .. (-0.05,1.8);
    \end{scope}
}

\newcommand{\butterfly}[1]{
    \begin{tikzpicture}

        \filldraw[fill=white, draw=none, line width=0pt] (0,0) circle (8.5cm);

        \draw[thick] (0,2.7) parabola[bend at end] (-1.5,5.2);
        \filldraw[fill=black!80!white!,draw=black,thick] (-1.5,5.2) .. controls ++(-0.2,0.1) and ++(-0.5,-0.3) .. ++(0,0);
        \begin{scope}[yscale=1,xscale=-1]
            \draw[thick] (0,2.7) parabola[bend at end] (-1.5,5.2);
            \filldraw[fill=black!80!white!,draw=black,thick] (-1.5,5.2) .. controls ++(-0.2,0.1) and ++(-0.5,-0.3) .. ++(0,0);
        \end{scope}

        \filldraw[fill=black!80!white!,draw=black,thick,rounded corners=2.5mm] (0,2.5) -- (0.35,2.5) -- (0.45,1.5) -- (0.45,0) -- (0.25,-2) -- (-0.25,-2) -- (-0.45,0) -- (-0.45,1.5) -- (-0.35,2.5) -- (0,2.5);
        \shade[inner color=black!70!white!,outer color=black!80!white!] (0,2.24) ellipse (0.35cm and 0.24cm);
        \shade[inner color=black!70!white!,outer color=black!80!white!] (0,1.75) ellipse (0.4cm and 0.25cm);
        \shade[inner color=black!70!white!,outer color=black!80!white!] (0,1.25) ellipse (0.42cm and 0.25cm);
        \shade[inner color=black!70!white!,outer color=black!80!white!] (0,0.75) ellipse (0.42cm and 0.25cm);
        \shade[inner color=black!70!white!,outer color=black!80!white!] (0,0.25) ellipse (0.42cm and 0.25cm);
        \shade[inner color=black!70!white!,outer color=black!80!white!] (0,-0.25) ellipse (0.4cm and 0.25cm);
        \shade[inner color=black!70!white!,outer color=black!80!white!] (0,-0.75) ellipse (0.35cm and 0.25cm);
        \shade[inner color=black!70!white!,outer color=black!80!white!] (0,-1.25) ellipse (0.3cm and 0.25cm);
        \shade[inner color=black!70!white!,outer color=black!80!white!] (0,-1.74) ellipse (0.25cm and 0.24cm);

        \shadedraw[inner color=black!60!white!,outer color=black!80!white!,draw=black,thick,rounded corners=2mm] (0,3) -- (0.45,3) -- (0.2,2.3) -- (-0.2,2.3) -- (-0.45,3) -- (0,3);
        \shadedraw[inner color=white!60!black,outer color=black, draw=black,thick] (-0.25,2.85) circle (0.2cm);
        \shadedraw[inner color=white!60!black,outer color=black, draw=black,thick] (0.25,2.85) circle (0.2cm);

        \begin{scope}[xshift=-0.35cm]
            \halfwing{#1}
        \end{scope}
        \begin{scope}[xshift=0.35cm]
            \halfwing{-#1}
        \end{scope}
    \end{tikzpicture}
}

\newcommand{\bffa}{\begin{animateinline}[autoplay,loop,scale=0.027]{13}
        \butterfly{1}
        \newframe
        \butterfly{0.883}
        \newframe
        \butterfly{0.6}
        \newframe
        \butterfly{0.317}
        \newframe
        \butterfly{0.2}
        \newframe
        \butterfly{0.317}
        \newframe
        \butterfly{0.6}
        \newframe
        \butterfly{0.883}
        \newframe
        \butterfly{1}
        \newframe
        \butterfly{0.883}
        \newframe
        \butterfly{0.6}
        \newframe
        \butterfly{0.317}
        \newframe
        \butterfly{0.2}
        \newframe
        \butterfly{0.317}
        \newframe
        \butterfly{0.6}
        \newframe
        \butterfly{0.883}
        \newframe
        \butterfly{1}
        \newframe
        \butterfly{0.883}
        \newframe
        \butterfly{0.6}
        \newframe
        \butterfly{0.317}
        \newframe
        \butterfly{0.2}
        \newframe
        \butterfly{0.317}
        \newframe
        \butterfly{0.6}
        \newframe
        \butterfly{0.883}
        \newframe[1.5]
        \butterfly{1}
    \end{animateinline}}

\newcommand{\ta}{\begin{animateinline}[autoplay,loop,poster=first]{12}
    \multiframe{60}{i=0+1}{
      \includegraphics[width=0.05\linewidth]{picture/tornado/frame_\i.png}
    }
\end{animateinline}
}

\renewcommand{\arraystretch}{0.95}

\title{\raisebox{-0.2em}{\bffa{}}\hspace{-3pt}\raisebox{-0.1em}{\ta{}} The Butterfly Effect of Model Editing:\\ Few Edits Can Trigger Large Language Models Collapse}

\author{Wanli Yang$^{\spadesuit}$, Fei Sun$^{\spadesuit}\footnotemark[2]$,
  Xinyu Ma$^{\clubsuit}$, Xun Liu$^{\heartsuit}$, Dawei Yin$^{\clubsuit}$, Xueqi Cheng$^{\spadesuit\heartsuit}$ \\
  $^{\spadesuit}$CAS Key Laboratory of AI Safety,\\ %
  Institute of Computing Technology, Chinese Academy of Sciences, Beijing, China \\
  $^{\heartsuit}$University of Chinese Academy of Sciences, Beijing, China ~~$^{\clubsuit}$Baidu Inc., Beijing, China \\
 \tt{yangyywl@gmail.com \,\, sunfei@ict.ac.cn}}

\begin{document}
\maketitle

\renewcommand*{\thefootnote}{\fnsymbol{footnote}}
\footnotetext[2]{Corresponding author.}
\renewcommand*{\thefootnote}{\arabic{footnote}}

\begin{abstract}

Although model editing has shown promise in revising knowledge in Large Language Models (LLMs), its impact on the inherent capabilities of LLMs is often overlooked.
In this work, we reveal a critical phenomenon: \textit{even a single edit can trigger model collapse}, manifesting as significant performance degradation in various benchmark tasks. %
However, benchmarking LLMs after each edit, while necessary to prevent such collapses, is impractically time-consuming and resource-intensive.
To mitigate this, we propose using perplexity as a surrogate metric, validated by extensive experiments demonstrating changes in an edited model's perplexity are strongly correlated with its downstream task performances.
We further conduct an in-depth study on sequential editing, a practical setting for real-world scenarios, across various editing methods and LLMs, focusing on hard cases from our previous single edit studies.
The results indicate that nearly all examined editing methods result in model collapse after only few edits. %
To facilitate further research, we have utilized GPT-3.5 to develop a new dataset, \textit{HardEdit}, based on those hard cases. %
This dataset aims to establish the foundation for pioneering research in reliable model editing and the mechanisms underlying editing-induced model collapse. %
We hope this work can draw the community's attention to the potential risks inherent in model editing practices\footnote{Code and data released at \url{https://github.com/WanliYoung/Collapse-in-Model-Editing}.}.

\if
Despite pre-trained language models demonstrate outstanding capabilities across various aspects, maintaining their knowledge correct and timely remains complex and resource-intensive. In recent years, Model Editing has rapidly evolved, aiming to modify models' knowledge by quickly adjusting their parameters. However, the impact of editing on the models remains unclear. In the exploration of side effects of Model Editing, we found some edit cases result in the models' collapse even with the use of state-of-the-art editing algorithms. To better evaluation, we empirically proved that perplexity of language models for human-written texts can verify if the models have been damaged. And we collect texts from commonly used pre-training corpora to build a dataset named \textbf{ME-PPL} for calculating perplexity. Further experiments indicate that the phenomenon of collapse is prevalent across different editing algorithms and language models in the setting of sequential editing on specific cases. For accessing whether Model Editing algorithms can ensure the models' stability in sequential editing scenario, we extend these cases to construct a hard to edit dataset named \textbf{Hard-COUNTERFACT}. Experiments reveal that most state-of-the-art editing algorithms result in model collapse with sequential editing on this dataset. 
\fi

\end{abstract}

\section{Introduction}

Large language models (LLMs) \cite{openai2023gpt4, touvron2023llama}, once trained, face the risk of becoming obsolete due to the dynamic nature of world knowledge. %
This challenge has spurred interest in \textit{model editing} %
\cite{yao2023editing}, an emerging research area dedicated to efficiently updating model parameters to modify outdated or incorrect knowledge in models, thus avoiding the huge costs of retraining from scratch \cite{meng2023locating}. 
Recently, model editing has advanced significantly and found applications in various domains, including question answering (QA) \cite{huang2023transformerpatcher}, hallucination correction \cite{hartvigsen2023aging}, and model repair \cite{murty2022fixing}. %

\begin{figure}[t]
    \begin{subfigure}{0.22\textwidth}
        \centering
        \includegraphics[width=\textwidth]{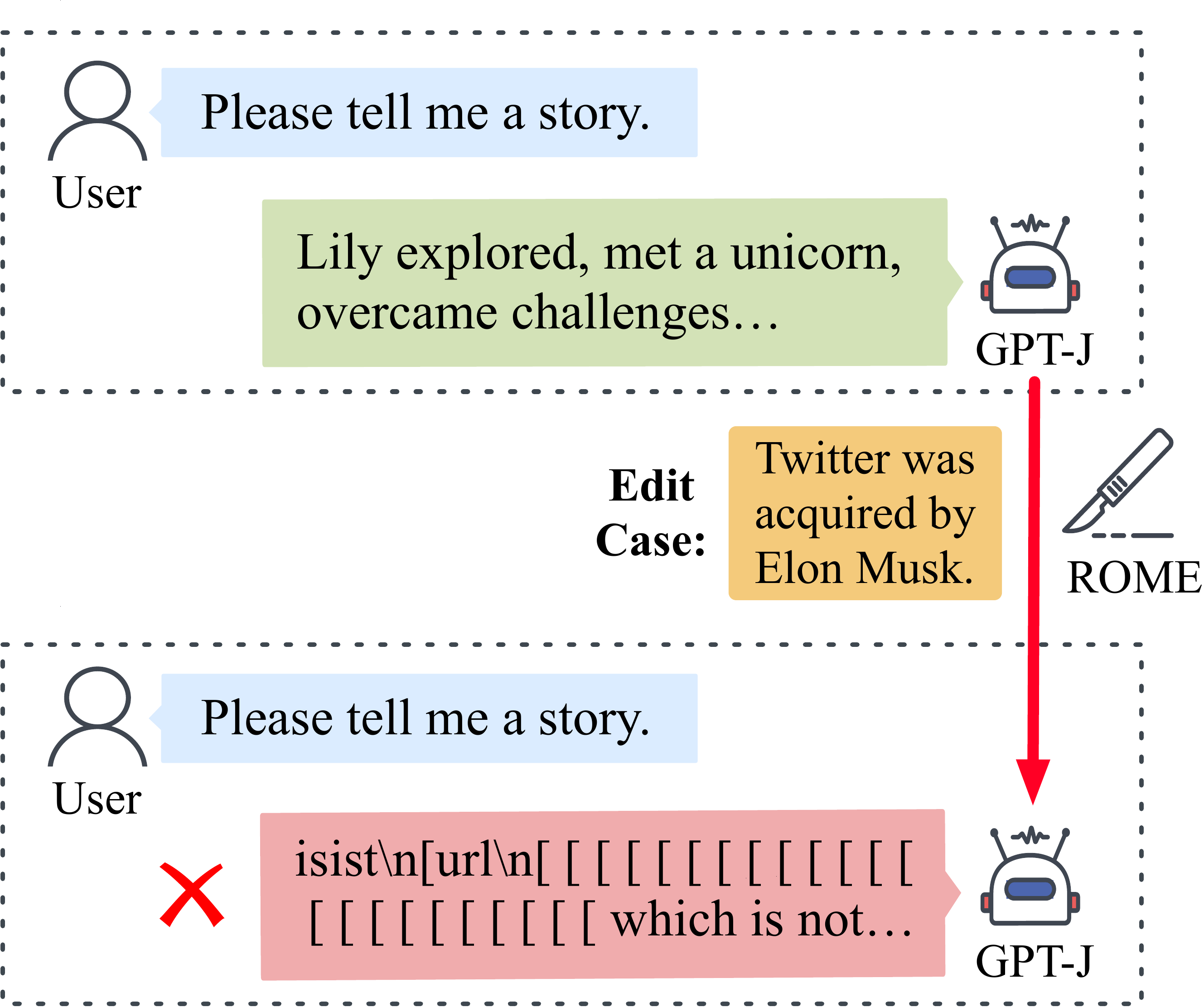}
        \captionsetup{skip=2pt}
        \caption{}
        \label{fig:collapse}
    \end{subfigure}
    \hfil
    \begin{subfigure}{0.24\textwidth}
        \centering
        \includegraphics[width=\textwidth]{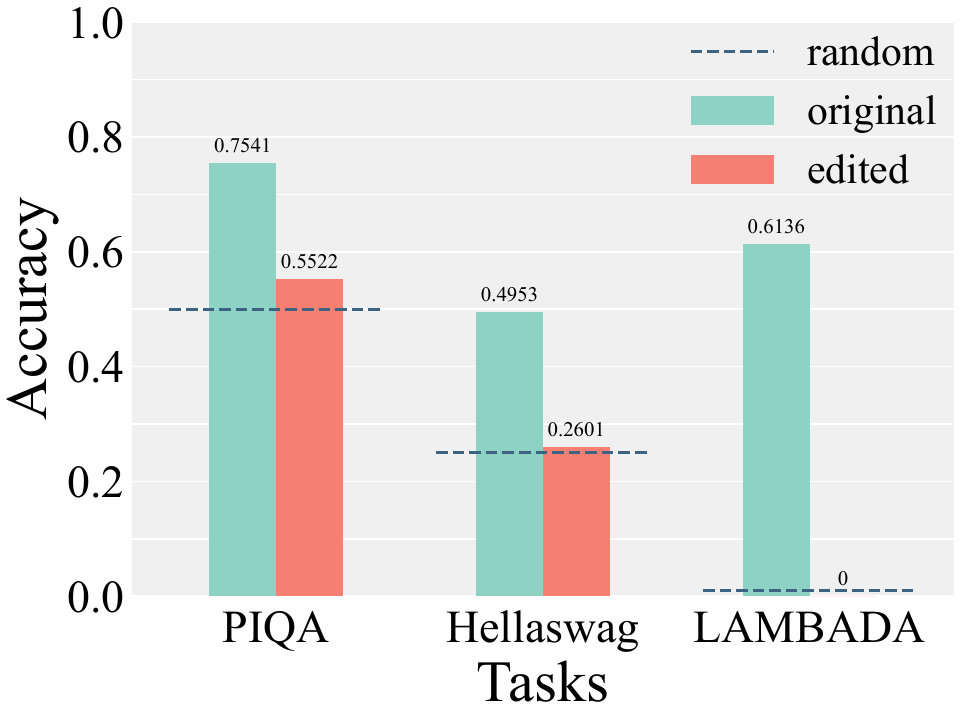}
        \captionsetup{skip=2pt}
        \caption{}
        \label{fig: collapse performance bar}
    \end{subfigure}
    \captionsetup{skip=0pt}
    \caption{(a) Editing GPT-J with ROME to inject a new fact ``Twitter was acquired by Elon Musk'' severely disrupts its ability to generate coherent text. (b) The downstream tasks performance of the edited GPT-J in Figure \ref{fig:collapse} has significantly deteriorated, approaching the ``random'' baseline indicative of mere guesswork.}
\end{figure}

However, our pilot explorations reveal a critical and unexpected risk: \textit{even a single edit can cause model collapse}.
As shown in Figure~\ref{fig:collapse}, employing ROME \cite{meng2023locating}, a cutting-edge model editing method, to update GPT-J with only one fact led to a marked deterioration in its text generation capabilities. %
Moreover, Figure~\ref{fig: collapse performance bar} highlights a significant decline in the performance of edited GPT-J on three representative tasks from its official evaluation task sets, approaching the level of random guessing on these tasks.
Herein, we term the phenomenon of significant performance decline across various downstream tasks in the edited model as ``\textit{model collapse}''.
This observation raises two critical questions for model editing:
\begin{itemize}[leftmargin=11pt, itemsep=3pt, topsep=3pt, parsep=2pt]
    \item \textit{How can we efficiently identify or measure collapse in an edited language model?}
    \item \textit{Is model collapse a common issue across different language models and editing methods?}
\end{itemize}

Although a thorough evaluation of edited models across downstream tasks for each edit offers a straightforward solution, the substantial time and resource consumption makes it impractical for real-world usage.
To streamline it, we propose using \textit{perplexity} to evaluate model collapse during editing and verify its efficacy in indicating downstream task performances through extensive experiments.
Furthermore, to ensure the reliability of perplexity computations, we curate a diverse and high-quality dataset \textbf{ME-PPL} (\textbf{M}odel \textbf{E}diting-\textbf{P}er\textbf{pl}exity) from various commonly used corpora.

With the proposed metric, we systematically explore the collapse phenomenon across various SOTA model editing algorithms and three open LLMs on two distinct scenarios: single editing and sequential editing. %
For \textit{single editing}, we reveal that applying ROME on the COUNTERFACT dataset leads to model collapse in all three LLMs under study.
Consequently, we gather samples that triggered model collapse in single edit trials, constituting the \textbf{HardCF} dataset, to streamline subsequent studies by focusing on the most problematic instances. %
For \textit{sequential editing}, a practical setting in real-world applications, we observe that model collapse occurs prevalently across almost all combinations of editing methods and LLMs we studied, within just dozens of edits on HardCF. %
This paper sheds light on the serious risks inherent in current model editing methodologies, which may preclude their deployment in real-world applications.

Inspired by the above findings, we build a challenging dataset called \textbf{HardEdit} to facilitate a more rigorous evaluation of the vulnerability of model editing algorithms to model collapse.
To populate this dataset with challenging examples, we utilize GPT-3.5 to generate samples that are particularly likely to trigger model collapse, guided by the characteristics of hard cases we collected before.
Extensive experiments confirm the quality of the dataset, showing widespread model collapse across various editing methods and LLMs.

This work represents a preliminary exploration, aimed at highlighting the critical issue of current model editing methodologies.
Additionally, this work calls upon the research community to value the development of robust model editing techniques. %
Our main contributions are as follows.
\begin{itemize}[leftmargin=11pt, itemsep=1pt, topsep=2pt]
    \item We unveil a hitherto unknown yet critical issue: a single edit can trigger model collapse.
    \item We propose to use perplexity for assessing the general capabilities of LLMs in model editing. %
    \item We demonstrate that model collapse is a ubiquitous issue for current editing algorithms in sequential edit setting via extensive experiments.
    \item We employ GPT-3.5 to construct a rigorous dataset HardEdit for enabling a comprehensive evaluation of model editing techniques, promoting further research and progress in the field.
\end{itemize}

\section{Background \& Study Formulation}

\subsection{Model Editing}

Model editing aims to modify a model's behavior on specific facts by directly adjusting its parameters instead of retraining, while preserving its behavior on irrelevant cases.
Formally, given an original fact $t {=} (s, r, o)$, consisting of subject $s$, relation $r$, and object $o$, encoded in an LLM $f_{\theta}$ and a revised fact $t' = (s, r, o')$ where $o' \neq o$, the objective of the editing algorithm $\xi$ is to optimize the parameter $\theta$ into $\theta'$ so that the edited model $f_{\theta'}\!\!:\! f_{\theta'} {=} \xi(f_{\theta}, t')$ correctly produces $o'$ when provided with the prompt $\mathtt{p}(s, r)$, as $f_{\theta'}(\mathtt{p}(s,r)) = o'$.
Using a presidential transition as an example, for the subject $s{=}$ \textit{United States} and relation $r{=}$ \textit{president of}, the editing algorithm $\xi$ ensures that the edited model $f_{\theta'}$ produces the expected object $o'{=}$ \textit{Joe Biden}, instead of previous $o{=}$ \textit{Donald Trump}, with prompt $\mathtt{p}(s, r)=$ \textit{The president of the United States is}.

\subsection{Current Methodologies}

Existing model editing methods can be broadly categorized into three groups. 

\noindent\textbf{Fine-tuning}. 
This intuitive paradigm mainly utilizes layer-wise fine-tuning to adjust parameters in light of new examples, simultaneously incorporating a constraint to ensure minimal interference with unmodified facts.
Typically, \citet{zhu2020modifying} propose fine-tuning LLMs within a norm constraint between edited and original model's parameters to mitigate the risk of catastrophic forgetting.
Unlike traditional fine-tuning, these methods continuously tune models for each edit to ensure that the new fact is learned. %

\noindent\textbf{Meta Learning}. 
Leveraging meta learning principles, this category of methods \cite{decao2021editing, mitchell2022fast, tan2023massive} usually employs a hypernetwork, serving as a helper model, to directly predict effective gradients or parameter modifications for encoding new facts.
\citet{decao2021editing} utilities a trained hypernetwork (a bidirectional-LSTM) to predict the parameters modification for each edit request.
\citet{mitchell2022fast} employs hypernetworks to learn a low-rank decomposition of the fine-tuning gradients to modify LLMs for new facts.
Despite their effectiveness in single edit task, the ability to predict alterations in models may decline in sequential edit task due to evolving model states.

\noindent\textbf{Locate-then-Edit}.
This paradigm is fundamentally grounded in the ``key-value memory'' hypothesis, positing that facts are encoded in the localized parameters of the transformer architecture, where the Feed-Forward Network (FFN) operates as key-value memory that supports factual association \cite{geva2021transformer}. 
Based on this, existing approaches attempt to localize target knowledge in specific parameters of models, and update these to inject new knowledge.
KN \cite{dai2022knowledge} employ knowledge attribution to identify the ``knowledge neuron'' (a key-value pair of FFN) which encodes certain knowledge, and then update the knowledge by modifying the neuron. 
ROME \cite{meng2023locating} utilizes causal tracing to localize knowledge at a specific MLP layer of a transformer, and then modify knowledge with rank-one update to the weight matrix.
MEMIT \cite{meng2023massediting} extends ROME by applying updates across multiple MLP layers, realizing massive edits.

\subsection{Evaluation of Edited Models}

The edited model $f_{\theta'}$ is typically evaluated from four properties: 
\begin{enumerate*}[label=\roman*)]
    \item \textit{reliability}, measuring the success rate of the edit;
    \item \textit{generalization}, evaluating the model's performance on equivalent edit prompts;
    \item \textit{locality}, examining the impact of the edit on irrelevant knowledge;
    \item \textit{portability}, assessing the model's performance on factual reasoning related to the editing request.
\end{enumerate*}
Interested readers are directed to \citet{yao2023editing} for an in-depth exploration.
Additionally, \citet{hoelscherobermaier2023detecting} claim a limitation in the currently used specificity (i.e., locality) metric, which focuses only on model responses to given prompts, and propose using KL divergence to measure changes in the full probability distribution of model outputs. 

\subsection{Side Effects of Model Editing}
Despite promising early results, the potential side effects of model editing have progressively garnered research interest as well.
\citet{yao2023editing} demonstrate that model editing algorithms may influence other relations associated with the subjects of edits, with the impact of FT$_{\bm{\ell_{\infty}}}$ \cite{zhu2020modifying} being particularly pronounced. 
\citet{hoelscherobermaier2023detecting} find that incorporating text relevant to edit cases into unrelated prompts can cause the responses of edited models to shift toward the target of the edits, which reveals that the models are over edited. 
\citet{brown2023edit} report that edits generally reduce the overall robustness of the model, and the degree of this reduction varies with the choice of editing algorithms and location.
Existing explorations of side effects primarily concentrate on the non-robust behaviors of model associated with editing.

\subsection{Research Question}

In this paper, we argue that for model editing to be practically useful, it is essential to ensure that the edited model maintains its abilities in downstream tasks.
Thus, we are interested in the following questions:
\begin{itemize}[leftmargin=11pt, itemsep=-4pt, topsep=0pt]
\item \textit{Can current model editing methods retain LLMs' inherent capabilities in downstream tasks?}
\item \textit{If not, how do current editing approaches affect LLMs' performance in real-world tasks?}
\item \textit{How can we efficiently identify or measure this impact for an edited language model?}
\end{itemize}
These are the main focus of our study, which will be discussed
in \S~\ref{sec:motivation}, \S~\ref{sec:ppl}, and \S~\ref{sec:study}.

\section{Experimental Setup}
\label{sec:ExpSetup}

This section outlines the basic setup of our study, serving as the default framework for all subsequent experiments unless otherwise noted.

\subsection{Editing Methods, Datasets, \& LLMs}

\noindent\textbf{\textit{Editing Methods}}.
For a comprehensive experimental scope, we employ four diverse and representative model editing methods from the three aforementioned categories:
fine-tuning (\textbf{FT}${_{\bm{\ell}_{\bm{\infty}}}}$, \citealp{zhu2020modifying}), meta-learning (\textbf{MEND}, \citealp{mitchell2022fast}), and locate-then-edit (\textbf{ROME}, \citealp{meng2023locating} and \textbf{MEMIT}, \citealp{meng2023massediting}). 
All these methods are implemented using \texttt{EasyEdit}\footnote{\url{https://github.com/zjunlp/EasyEdit}}. 
For the training-required method, MEND, the split of datasets follows the common practice as in \cite{decao2021editing, mitchell2022fast}.

\noindent\textbf{\textit{Editing Datasets}}. We employ the two most prevalent benchmark datasets:
\textbf{ZsRE} \cite{levy2017zero} and \textbf{COUNTERFACT} \cite{meng2023locating}.
For ZsRE, we adopt the established data split from \cite{meng2023locating, yao2023editing}, using the test set (\num{10000} records) for our study.

\noindent\textbf{\textit{Backbone LLMs}}.
Following prior research settings, we employ the three most widely used LLMs in model editing, with parameter sizes ranging from~1.5 to~7 billion to reflect a diverse set of capabilities: \textbf{GPT-2-XL} (1.5 billion parameters)  \cite{radford2019language}, \textbf{GPT-J} (GPT-3-like LLM with~6 billion parameters) \cite{gpt-j}, and \textbf{Llama2-7b} (a leading open-source LLM with~7 billion parameters) \cite{touvron2023llama}.
For all the LLMs under investigation, greedy decoding is consistently adopted during text generation and downstream task evaluation.

\subsection{Representative Tasks}
\label{subsec: tasks}

To assess the overall capabilities of the edited models, we choose six representative tasks from the collective set of official evaluation benchmarks for the LLMs under study.
Our evaluation encompasses two categories, each with three tasks, to probe distinct capabilities of the model: Hellaswag \cite{zellers2019hellaswag}, PIQA \cite{Bisk2020}, and MMLU \cite{hendrycks2021mmlu} for discriminative abilities; and LAMBADA \cite{paperno-EtAl:2016:P16-1}, Natural Questions (NQ) \cite{kwiatkowski2019nq}, and SQuAD2.0 \cite{rajpurkar2018squad} for generative capacities.
Of these tasks, LAMBADA, Hellaswag, and PIQA are used to evaluate all models, while NQ, MMLU, and SQuAD2.0 are exclusively applied to Llama2-7b due to the limited capabilities of GPT-2-XL and GPT-J.
For efficiency, we select 4 out of the 57 subtasks of MMLU to form MMLU$_{\mathit{sub}}$, which effectively represents its core categories, for subsequent study.
Evaluation of these tasks is performed using \texttt{lm-eval} package\footnote{\url{https://github.com/EleutherAI/lm-evaluation-harness}}.

Further descriptions of the methods, datasets, models, and tasks can be found in Appendix~\ref{appendix:ExpSet}.

\section{Pilot Observation}
\label{sec:motivation}

This section introduces the motivation of our research, a pilot exploration to elucidate the side effects of model editing on LLMs.

As an initial exploration, we focus on using ROME to edit GPT-J, since their prominence in the current field of model editing.
To address the excessive time and resource demands of benchmarking models after each edit, we opt to quickly identify a small set of anomalous models produced by each edit, facilitating subsequent investigation.
Inspired by recent studies linking perplexity with linguistic competence in LLMs \cite{zhao2023unveiling}, we initially employ perplexity as a tool to detect such anomalies.
For computational efficiency, we utilize a subset of \num{50} sentences from the dataset in \S~\ref{sec:ppl} to expedite the perplexity calculations.
A comprehensive examination of perplexity as a metric for assessing model collapse is presented in \S~\ref{sec:ppl}.

\begin{figure}[t]
    \begin{subfigure}{0.23\textwidth}
        \centering
        \includegraphics[width=\textwidth]{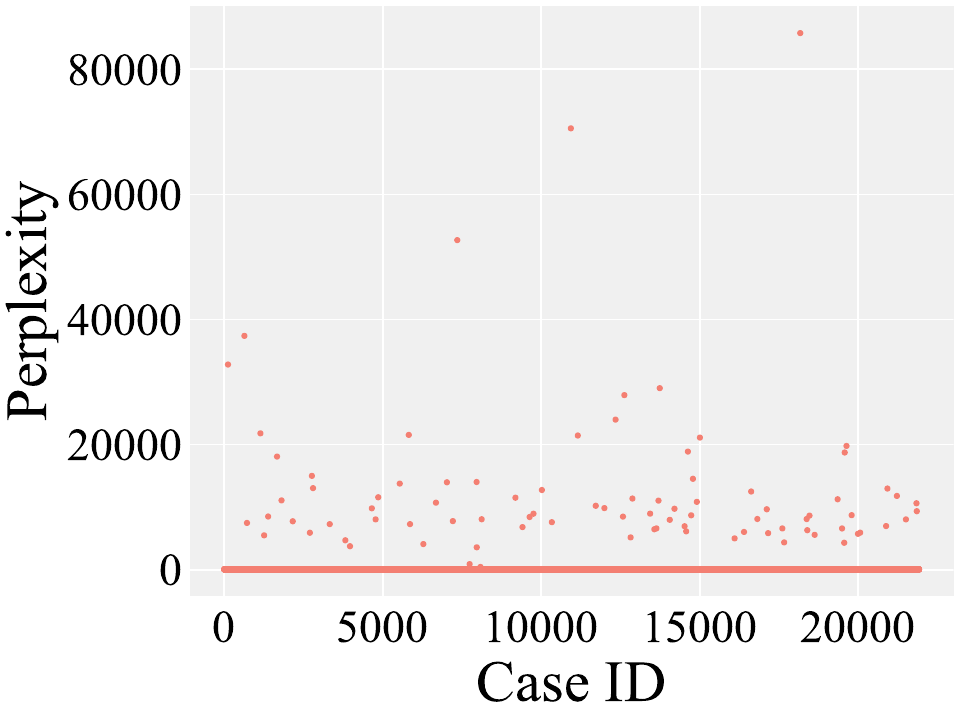}
        \caption{}
        \label{fig:rome_gptj_ppl}
    \end{subfigure}
    \begin{subfigure}{0.23\textwidth}
        \centering
        \includegraphics[width=\textwidth]{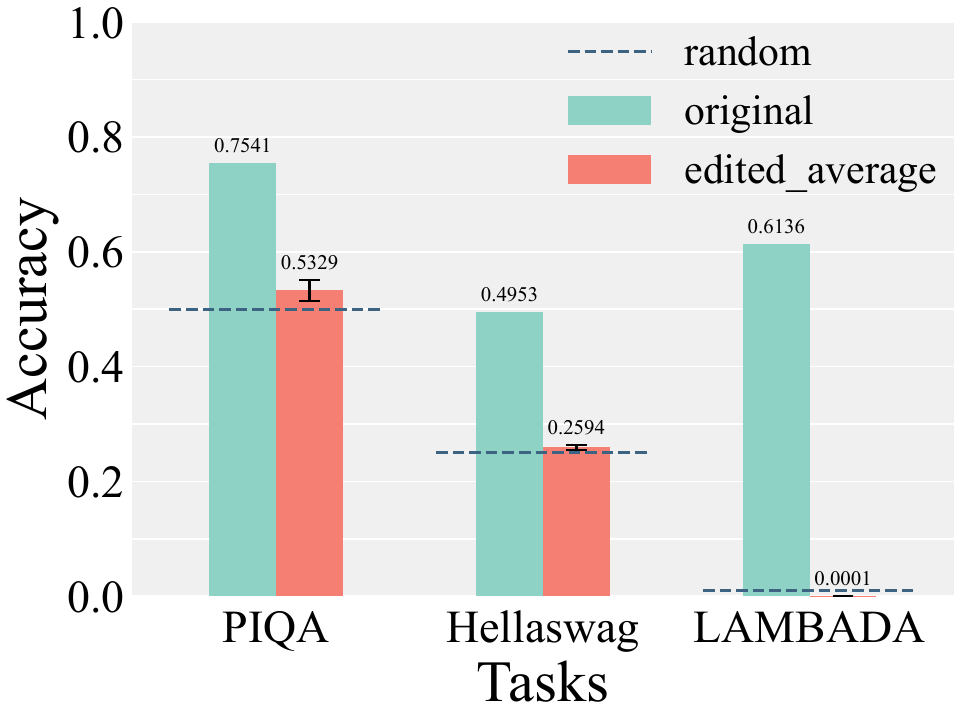}
        \caption{}
        \label{fig:top30_perform}
    \end{subfigure}
    \caption{(a) Scatter plot of perplexity for models independently edited by ROME from the original GPT-J, with each point representing a unique edit case in the COUNTERFACT dataset. ``Case ID'' refers to the index of each edit sample. (b) Average performance with variance on downstream tasks for the top 30 high-perplexity models in Figure~\ref{fig:rome_gptj_ppl}, comparing to the original model and random guessing.}
\end{figure}

Figure~\ref{fig:rome_gptj_ppl} illustrates the results of employing ROME to edit GPT-J on the COUNTERFACT dataset with single edit setting.
For brevity, the results of ZsRE, which show no anomalies, are detailed in Appendix~\ref{appendix:zsre}.
Each point in the figure represents the perplexity of a model edited independently from the original GPT-J, using a unique sample from the COUNTERFACT dataset.
Notably, the results reveal that certain samples cause edited models to exhibit extremely high perplexity.

To understand what occurred in these cases, we chose the edited models with top \num{30} highest perplexity in Figure~\ref{fig:rome_gptj_ppl}, and evaluated their performances on the discrimination tasks (PIQA and Hellaswag) and the generation task (LAMBADA). 
All the models' performances markedly decline on these downstream tasks as shown in Figure~\ref{fig:top30_perform}. 
A subsequent basic text generation test with a high perplexity model confirmed the severity of the issue: the model lost its ability to generate coherent text, generating meaningless content instead, as shown in Figure~\ref{fig:collapse}.

Arising from this preliminary investigation, we uncover a previously unreported phenomenon that model editing can precipitate model collapse.
Naturally, this finding leads to two key questions: 
\begin{itemize}[leftmargin=11pt, itemsep=3pt, topsep=0pt, partopsep=0pt, parsep=0pt]
    \item Can perplexity effectively signal collapses in edited models, i.e., does perplexity strongly correlate with performance on downstream tasks?
    \item Is model collapse a common issue across various language models and editing methods?
\end{itemize}

\section{Perplexity as a Surrogate Metric}
\label{sec:ppl}

In this section, we conduct an in-depth investigation to assess whether perplexity can serve as a surrogate metric, closely correlating with downstream tasks performance, thereby avoiding the need for costly benchmarking LLMs after each edit.

Perplexity \cite{brown-etal-1992-ppl} is a conventional metric for measuring the generative capability of language models, defined as the exponential of the average negative log-likelihood of a sequence.
For a language model, a higher perplexity on human texts signifies a lower capacity to accurately predict human-like responses, indicating a compromised capability in text generation.
Furthermore, from a theoretical perspective, perplexity's exponential relationship with the training loss of LLMs \cite{radford2018loss} establishes it as a surrogate metric for assessing the status of the model.

\noindent\textbf{\textit{Dataset}}.
Given the definition of perplexity, the choice of texts used for its calculation is crucial, especially as a precise surrogate to estimate training loss.
Thus we construct the ME-PPL (\textbf{M}odel \textbf{E}diting-\textbf{P}er\textbf{pl}exity) dataset, comprised of \num{10000} uniformly lengthed, English sentences that are randomly sampled and processed from widely used corpora, e.g., BookCorpus \cite{zhu2015aligning}, Wikipedia \cite{wikidump}, and OpenWebText \cite{Gokaslan2019OpenWeb}.
To facilitate perplexity calculation in various situations, e.g, different computational load, we create two subsets, ME-PPL$_{50}$ with \num{50} sentences and ME-PPL$_{\text{1k}}$ with \num{1000} sentences. 
More details can be seen in Appendix~\ref{appendix: ppl}.
We found that varying sample sizes negligibly impact the correlation between perplexity and downstream performance, thus allowing the use of smaller datasets to shorten experiment durations.
In this section, we adopt ME-PPL$_{\text{1k}}$ for a more precise investigation.

\noindent\textbf{\textit{Experimental Setup}}.
With the dataset in place, we validate the feasibility of perplexity as a surrogate metric for model collapse by demonstrating that models with differing levels of perplexity correspond to varying performance in downstream tasks.
For this purpose, we apply model editing to establish a comprehensive range of perplexity levels, including twenty points distributed as uniformly as possible between the perplexity of original model and a threshold of \num{1000}, along with three additional points beyond this (specifically, $5{\times}10^3$, $1{\times}10^4$, and $5{\times}10^4$) to represent collapsed models.
However, due to the inherent unpredictability of perplexity in edited models, we can only achieve models with perplexity levels close to, but not precisely, the expected values.

\begin{figure}[t]
    \centering
    \includegraphics[width=0.49\textwidth]{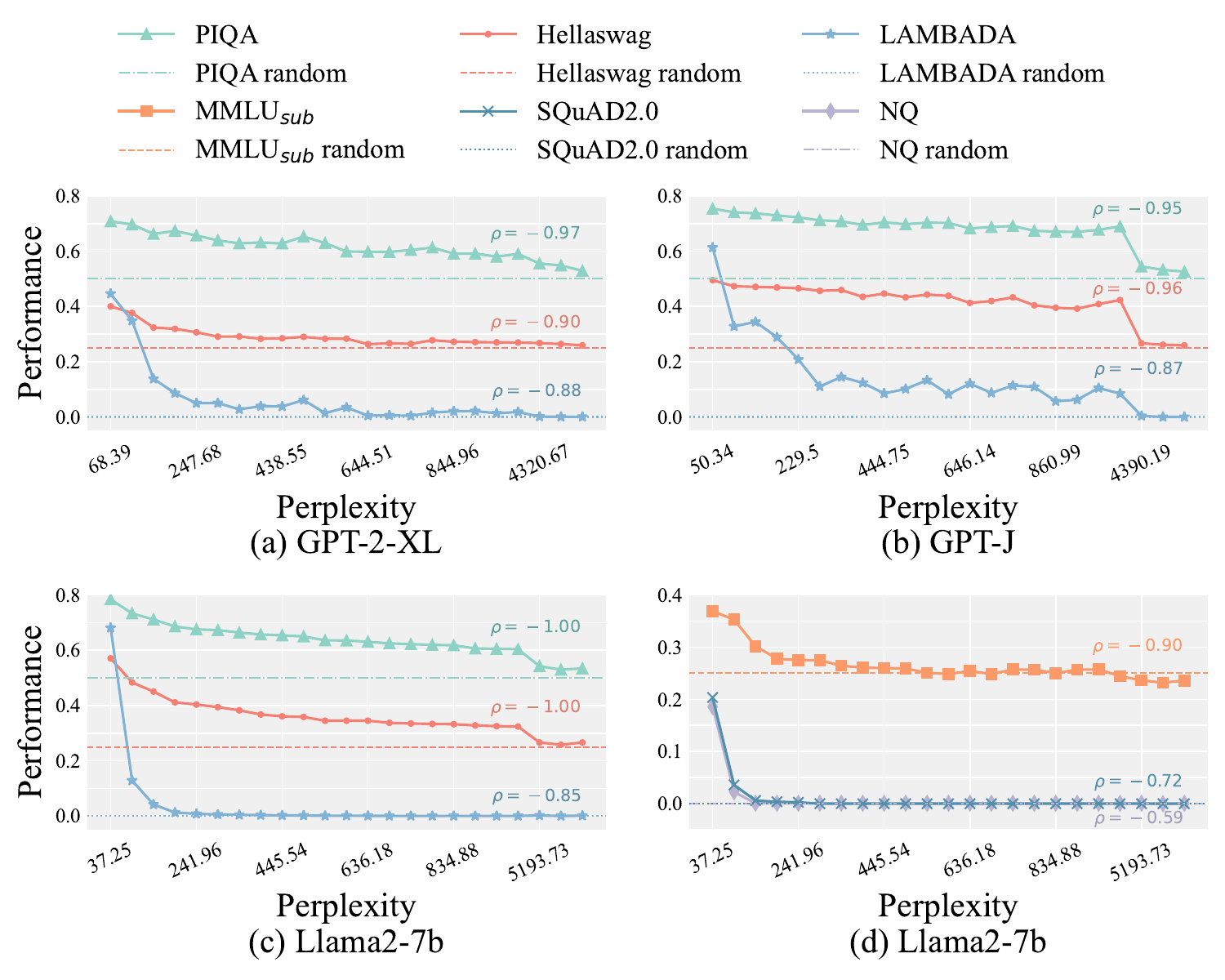}
    \caption{Correlations between perplexity and downstream task performance across different LLMs, measured by task-specific metrics: Exact Match (EM) for NQ; F$_1$ for SQuAD2.0.; Accuracy for remaining tasks.
    $\rho$ refers to the Spearman's Rho value, measuring the rank correlation between perplexity and corresponding downstream task performance, with all $p$-values $<$ \num{0.01}.}
    \label{fig_ppl4eval}
\end{figure}

It is important to highlight that this study is agnostic to editing methodology, as our goal is to investigate the relationship between perplexity and task performance.
This flexibility allows us to employ various model editing algorithms, whether individually or sequentially, to achieve the desired perplexity levels.
For example, we successfully got a Llama2-7b model to reach a perplexity of \num{9613.17} (roughly \num{10000}) by applying a single edit via ROME.
Conversely, by sequentially applying FT$_{\ell_{\infty}}$ \num{18} times, we obtained a Llama2-7b model with a perplexity of \num{97.25} (around \num{100}).
Finally, we obtained models with 23 distinct perplexity variations for each of the three models and subsequently evaluated these edited models on the tasks introduced in \S~\ref{sec:ExpSetup}.

\noindent\textbf{\textit{Results}}.
The results in Figure~\ref{fig_ppl4eval} reveal a significant correlation between the perplexity of LLMs and their performance on downstream tasks. 
Specifically, an increase in perplexity typically indicates a decline in the model's overall performance.
It is noteworthy that the lower $\rho$ values for NQ and SQuAD2.0 are attributed to the premature decline in task performance to a level of random guessing.
Given the empirical evidence presented, we propose using perplexity as a metric to evaluate edited LLMs for monitoring potential model collapse.
It is essential to emphasize that our intention is to employ perplexity to monitor the drastic change in an edited model, rather than as a precise measure of comparative capabilities across various LLMs as in \citet{hu2024can}.

\subsection{Discussion}

Additionally, there are other metrics designed to assess the side effects of model editing, e.g., locality \cite{yao2023editing}, as well as consistency and fluency \cite{meng2023locating}.
However, these metrics are insufficiently effective, especially in detecting model collapse.
In this section, we discuss the connections and differences between perplexity and these metrics.

\begin{table}[t]
    \centering
    \begin{adjustbox}{max width=\linewidth} 
    \begin{tabular}{lrrrrr}
        \toprule
        Edit Case & locality $\uparrow$ & RS $\uparrow$ & GE $\uparrow$ & perplexity $\downarrow$ & PIQA $\uparrow$ \\
        \midrule
        \parbox[c]{4.5cm}{Motion, a product manufactured by \sethlcolor{mine_green}\hl{Apple} $\rightarrow$ \sethlcolor{pink}\hl{Microsoft}} & \num{1} & \num{69.08} & \num{612.29} & \num{6274.74} & \num{0.5462} \\
        \midrule
        \parbox[c]{4.5cm}{Vanderbilt University, whose headquarters are in \sethlcolor{mine_green}\hl{Nashville} $\rightarrow$ \sethlcolor{pink}\hl{Toronto}} & \num{0} & \num{43.65} & \num{642.65} & \num{68.38} & \num{0.7078}	 \\
        \bottomrule
    \end{tabular}
    \end{adjustbox}
    \caption{Comparison between perplexity and existing metrics, locality, consistency (\textbf{R}eference \textbf{S}core), and fluency (n\textbf{G}ram \textbf{E}ntropy), in assessing the edited GPT-2-XL's capabilities, using PIQA as the benchmark. 
    The computations for RS and GE are based on the code of ROME \cite{meng2023locating}.}
    \label{tab:locVsppl}
\end{table}

\noindent\textbf{\textit{Locality}}.
It evaluates the side effects of editing algorithms by examining whether the edited model changes its outputs on randomly sampled, irrelevant questions \cite{meng2023locating,yao2023editing}.
However, it often falls short as a comprehensive evaluation metric due to its limitations: insufficient sampling volume to cover all potential out-of-scope scenarios and the trivial nature of the employed token completion task that fails to capture the full range of LLM functionalities.
Table~\ref{tab:locVsppl} highlights the inconsistency of locality in practical usage, indicating model collapse at a value of \num{1} and stability at \num{0}, which contradicts actual model performance.

\noindent\textbf{\textit{Consistency and Fluency}}.
\citet{meng2023locating} assess the generative capabilities of edited models through consistency and fluency.
Consistency measures the cosine similarity between the generated texts and given reference texts while fluency focuses on identifying repetitive word patterns via bi- and tri-gram entropies.
However, a collapsed model may still produce texts with low repetition, while texts generated by a stable model might significantly diverge from the reference texts, as shown in Table~\ref{tab:locVsppl}.
This reveals the inadequacies of consistency and fluency as indicators of a model's generative capabilities.

The failure of previous works to identify model collapse may stem from their evaluation on sampled test data or basing their analysis on average metrics, resulting in the oversight of a small fraction of collapsed samples.

\section{Model Collapse Induced by Editing}
\label{sec:study}

This section is dedicated to using perplexity to systematically investigate collapse induced by model editing in single and sequential editing scenarios.

\subsection{Single Editing}
\label{Sing_Col}

Single editing is the fundamental and prevalent experiment setting in model editing research.
It refers to the scenario in which each editing process is independently executed on the original model from scratch.
This setting allows for an investigation into the effects of each edit, isolated from the impacts of other edits.

\noindent\textbf{\textit{Experiment Setup}}.
We conduct experiments using four editing methods\footnote{For the less effective editing method, KN, the results of single editing are provided in Appendix~\ref{apd_kn}, highlighting the frequent occurrence of edited model collapse.} on three LLMs across two datasets, as detailed in \S~\ref{sec:ExpSetup}.
Given the significant time for \num{24} ($3{\times} 4 {\times} 2$) different experimental setups, each requires tens of thousands of evaluations, we opted for ME-PPL$_{50}$ to accelerate perplexity calculation.
As shown in Figure~\ref{fig_ppl4eval}, a perplexity threshold of \num{1000} is employed to identify model collapse.

\subsubsection{Results \& Analysis}
\label{analysis of hard cases}

Upon examining the perplexity, we find that model collapse caused by a single edit exists in all three LLMs when applying ROME to COUNTERFACT.
Due to space limitations, we present the perplexity results for various experimental settings in Appendix~\ref{appendix:complete_ppl}.
Within COUNTERFACT, collapses were induced in \num{77} instances by GPT-2-XL, \num{85} by GPT-J, and \num{21} by Llama2-7b, respectively.
To facilitate subsequent studies, we aggregate these instances into a challenging subset named \textit{HardCF}, comprising \num{107} unique samples.

\noindent\textbf{\textit{Characteristics of HardCF}}. Table~\ref{tab_hard_case} presents some cases of HardCF, with additional cases elaborated in Appendix \ref{apd_hard_case}. 
For GPT-2-XL and GPT-J, the samples causing model collapse exhibit a high degree of overlap, primarily featuring subjects that are single, commonly used words, and are positioned at the beginning of the prompts.
For Llama2-7b, the subjects in these challenging cases usually encompass names of individuals or entities, ending with a period ``.''. %

\begin{table}[t]
    \centering
    \begin{adjustbox}{max width=\linewidth} 
    \begin{tabular}{lr@{\hskip 4pt}c@{\hskip 4pt}ll}
        \toprule
        \textbf{Model} & \multicolumn{3}{c}{\textbf{Edit Case}} \\
        \midrule
        \multirow{3}{*}{GPT-2-XL}  & \sethlcolor{pink}\hl{Arthur} is \uline{located} in \uwave{Illinois} & $\longrightarrow$ & \uwave{California} \\
        & \sethlcolor{pink}\hl{Q} was originally \uline{aired} on \uwave{BBC} & $\longrightarrow$ & \uwave{NBC} \\
        & \sethlcolor{pink}\hl{Minecraft}, \uline{created} by \uwave{Microsoft} & $\longrightarrow$ & \uwave{IBM} \\
        \midrule
        \multirow{3}{*}{GPT-J}    & \sethlcolor{pink}\hl{Flickr} \uline{owner} \uwave{Yahoo} & $\longrightarrow$ & \uwave{Houston} \\
                  & \sethlcolor{pink}\hl{Canada} is \uline{a part} of the \uwave{NATO} & $\longrightarrow$ & \uwave{FIFA} \\
                  & \sethlcolor{pink}\hl{Revolution} \uline{premieres} on \uwave{NBC} & $\longrightarrow$ & \uwave{HBO} \\
        \midrule
        \multirow{3}{*}{Llama2-7b} & \sethlcolor{pink}\hl{Call Cobbs, Jr.} \uline{performs} \uwave{jazz} & $\longrightarrow$ & \uwave{fantasy} \\
        & \sethlcolor{pink}\hl{Joe Garagiola Sr.} \uline{plays} \uwave{baseball} & $\longrightarrow$ & \uwave{hockey} \\
        & \sethlcolor{pink}\hl{Clint Murchison, Jr.} is \uline{native} to \uwave{Dallas} & $\longrightarrow$ & \uwave{Lyon} \\
        \midrule
        \multirow{3}{*}{Normal} & \sethlcolor{pink}\hl{Jon Larsen} \uline{plays} \uwave{jazz} & $\longrightarrow$ & \uwave{opera} \\
        & \sethlcolor{pink}\hl{Alexander VIII} \uline{expired at} \uwave{Rome} & $\longrightarrow$ & \uwave{London} \\
        & \sethlcolor{pink}\hl{Laurie Anderson} \uline{works as} \uwave{poet} & $\longrightarrow$ & \uwave{actor} \\
        \bottomrule
    \end{tabular}
    \end{adjustbox}
    \caption{Examples of HardCF that induce collapse in corresponding LLMs through a single ROME edit, with the ``Normal'' row showcasing other normal cases from COUNTERFACT for contrast.}
    \label{tab_hard_case}
\end{table}

{
\begin{table}[t]
    \centering
    \begin{adjustbox}{max width=\linewidth} 
    \begin{tabular}{llcccr}
        \toprule
        Model & Status & PIQA & Hellaswag & LAMBADA & perplexity \\

        \midrule
          & random & \num{0.5000} & \num{0.2500} & \num{0.0000} & -- \\

        \midrule
        \multirow{2}{*}{GPT-2-XL} & original & \num{0.7084} & \num{0.4004} & \num{0.4461} & \num{68.39} \\
         & edited & \num{0.5272} & \num{0.2568} & \num{0.0000} & \num{179837.93}\\

        \midrule
        \multirow{2}{*}{GPT-J} & original & \num{0.7541} & \num{0.4953} & \num{0.6136} &	\num{50.34} \\
         & edited & \num{0.5185} & \num{0.2617} & \num{0.0000} & \num{184391.46} \\
         
        \midrule
        \multirow{2}{*}{Llama2-7b} & original & \num{0.7845} & \num{0.5706} & \num{0.6814} & \num{37.25} \\
         & edited & \num{0.5087} & \num{0.2610} & \num{0.0008} & \num{7751.07} \\
         
        \bottomrule
    \end{tabular}
    \end{adjustbox}
    \caption{Performance comparison of highest-perplexity edited models  against the original models across various tasks, with ``random'' row denoting random guessing.}
    \label{tab_collapse}
\end{table}
}

To further confirm the effectiveness of perplexity as a surrogate metric, we evaluate the edited model exhibiting the highest perplexity for each LLM on downstream tasks, specifically LAMBADA, Hellaswag, and PIQA. 
Table \ref{tab_collapse} demonstrates that these models are severely damaged, 
further supporting the finding that a single edit can disrupt LLMs.

\begin{figure}[t]
    \centering
    \includegraphics[width=0.49\textwidth]{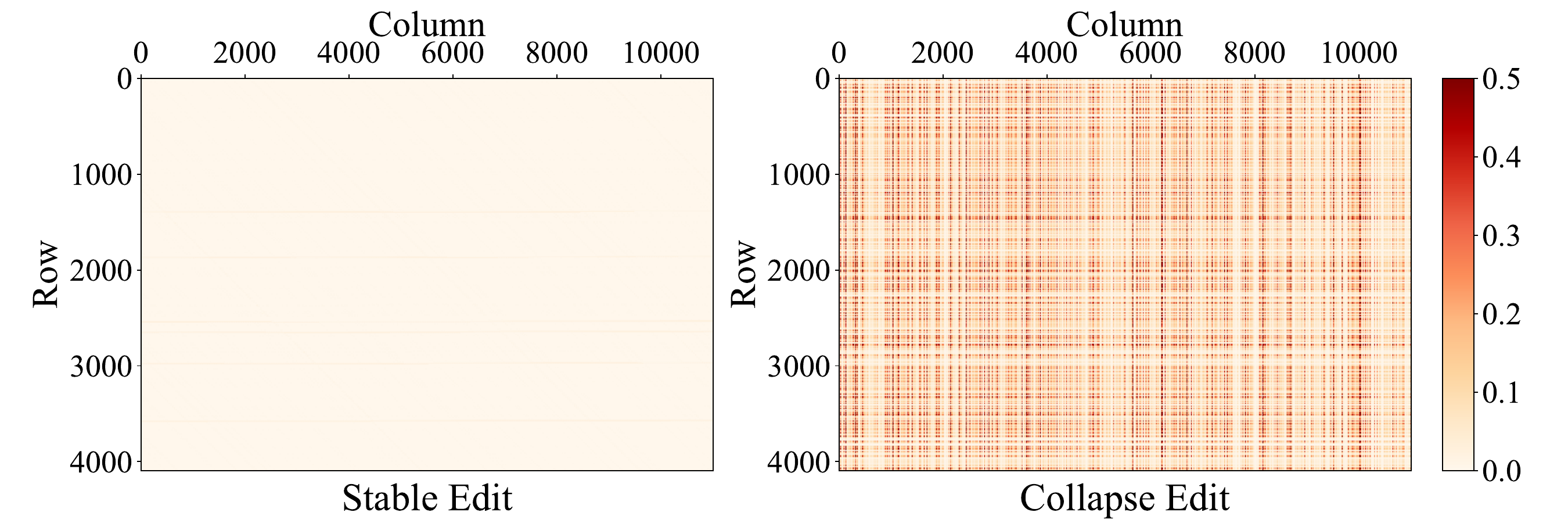}
    \caption{The absolute difference between the weights of the edited layer (Layers.5.mlp.down\_proj) and its original weights for ROME-edited Llama2-7b models.}
    \label{fig:heatmap}
\end{figure}

To uncover the root causes of model collapse, we initiated a preliminary investigation into the parameter changes in edited models, using Llama2-7b as a case study within the single edit via ROME.
We selected an edited model with the highest perplexity of \num{7751.07} as previously mentioned and another randomly sampled stable edited model with a perplexity of \num{37.25}, for comparison.
Figure~\ref{fig:heatmap} illustrates the absolute value of weight changes in the edited layer for each edit.
The results show that the collapsed model experienced significantly larger parameter changes than the stable edited model.

\begin{figure*}[t]
    \centering
    \includegraphics[width=\textwidth]{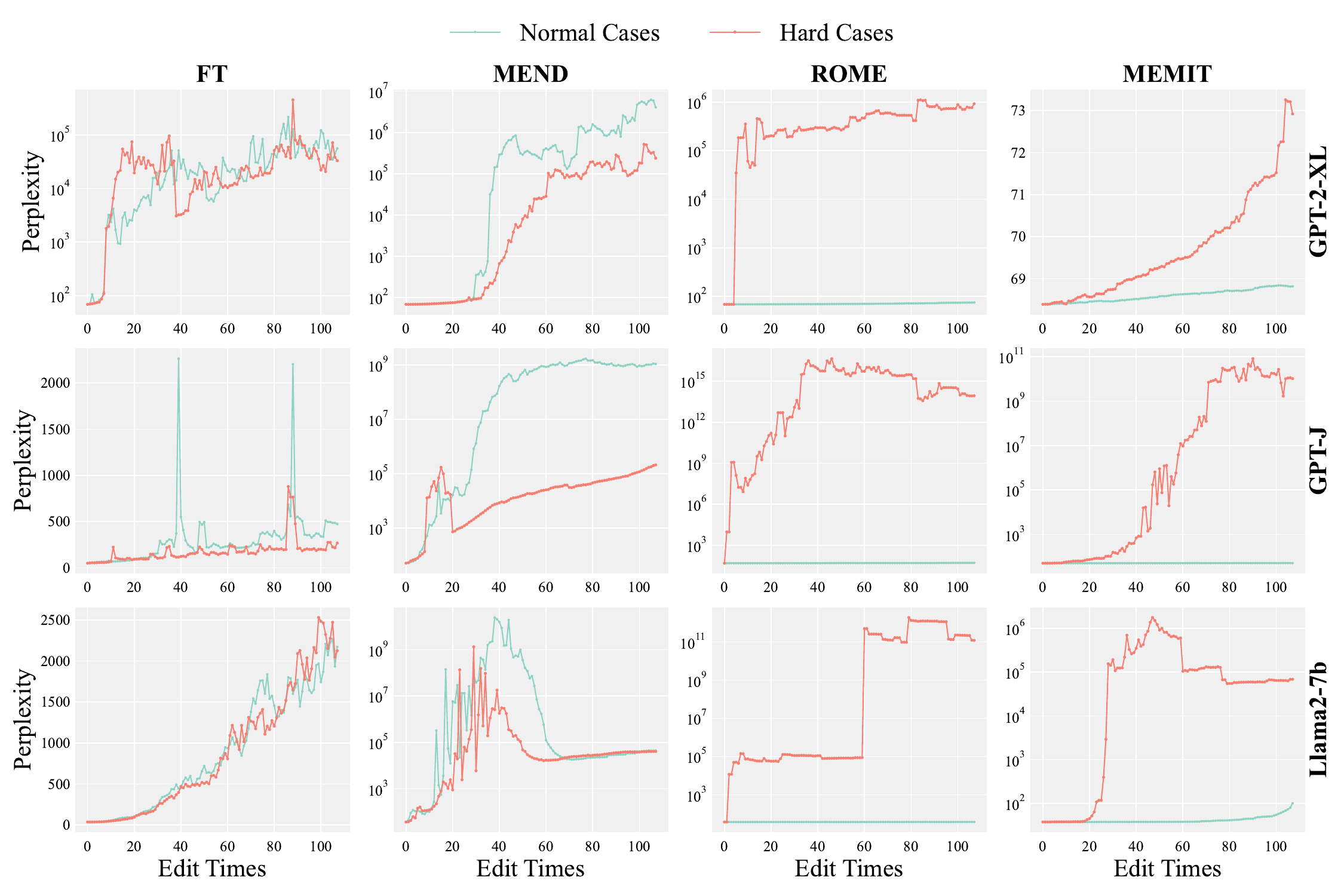}
    \captionsetup{skip=-2pt}
    \caption{Perplexity evolution over 107 editing iterations for normal and hard cases. 
    The y-axes are tailored for each subplot accordingly due to the the significant variation in the magnitude of perplexity changes.} %
    \label{fig_seq_collapse}
\end{figure*}

\subsection{Sequential Editing}
\label{sec:seq_edit}

Unlike single editing, which focuses on the impact of an individual edit, sequential editing is essential for the continuous knowledge updates in real-world applications.
It involves performing a series of edits in succession, with each subsequent edit meticulously crafted to preserve the integrity of previous edits \cite{huang2023transformerpatcher}.
Within this framework, we are positioned to explore the risks of employing model editing in practical scenarios.

\noindent\textbf{\textit{Experiment Setup}}.
We conduct a comparative study of the behaviors and risks of the editing algorithms in both hard and normal samples: \num{107} hard instances of HardCF and an equal number of normal samples randomly selected from the rest of COUNTERFACT.
We then execute sequential edits on each group separately, encompassing four editing algorithms and three LLMs\footnote{Experiments on Mistral-7b are also conducted, exhibiting phenomena akin to those of the three LLMs. The results are detailed in Appendix~\ref{apd_mistral}.} as in single edit experiments.
Notably, in light of the relatively small number of edits required for this experiment, the corpus for perplexity computation is expanded to ME-PPL$_{\text{1k}}$ for more precision.

\subsubsection{Results \& Analysis}
\label{subsec:seq_res}

The results of the sequential editing evaluation across various editing methods and LLMs are presented in Figure \ref{fig_seq_collapse}.
It can be observed that:

Nearly all editing methods caused model collapse during sequential editing on hard data, with the collapse occurring in remarkably few times---less than \num{60}.
The exception within this study was MEMIT applied to GPT-2-XL, and FT$_{\bm{\ell_{\infty}}}$ to GPT-J.
Further analysis reveals that although MEMIT avoided collapse (final perplexity of \num{72.92}), it edits successfully only in \num{23} out of \num{107} attempts, indicating very limited efficacy in model editing.
While FT$_{\bm{\ell_{\infty}}}$ did not induce total collapse in GPT-J, it significantly increased perplexity exceeding fivefold (from \num{50.34} to \num{268.61}) and impaired downstream task performance according to Figure~\ref{fig_ppl4eval}.

Another observation is the two distinct patterns in the four editing methods when applied to hard versus normal samples:
\begin{enumerate*}[label=\roman*)]
    \item FT$_{\bm{\ell_{\infty}}}$ and MEND behave similarly on both hard and normal samples, leading to their failure under each condition.
    \item In contrast, ROME and MEMIT exhibit significantly greater robustness, collapsing only in hard samples while maintaining stable perplexity in normal samples. 
\end{enumerate*}
This marked difference highlights the superiority of ROME and MEMIT, yet they still fall short of handling sequential edits on hard samples.

\begin{table}[t]
    \centering
    \renewcommand{\arraystretch}{1.05}
    \begin{adjustbox}{max width=\linewidth} 
    \begin{tabular}{lrcccccc}
        \toprule
        Method & perplexity & PIQA & Hellaswag & MMLU$_{\mathit{sub}}$ & LAMBADA & NQ & SQuAD2.0  \\
        
        \midrule
        original  & \num{37.25} & \num{0.7845} & \num{0.5706} & \num{0.3691} & \num{0.6814} & \num{0.1859} & \num{0.2036}  \\
        random  & -- & \num{0.5000} & \num{0.2500} & \num{0.2500} & \num{0.0000} & \num{0.0000} & \num{0.0000} \\
        \midrule
        \multicolumn{8}{c}{Normal Cases} \\
        \midrule
        
        FT$_{\bm{\ell_{\infty}}}$ & $2.17 \times 10^{3}$ & \num{0.5762} & \num{0.2990} & \num{0.2770} & \num{0.0002} & \num{0.0000} & \num{0.0003}  \\

        MEND & $4.46 \times 10^{4}$ & \num{0.5158} & \num{0.2546} & \num{0.2561} & \num{0.0000} & \num{0.0000} & \num{0.0003} \\
         
        ROME & $3.75 \times 10^{1}$ & \num{0.7797} & \num{0.5659} & \num{0.3681} & \num{0.6726} & \num{0.1731} & \num{0.1894} \\

        MEMIT & $9.98 \times 10^{1}$ & \num{0.7067} & \num{0.4749} & \num{0.2834} & \num{0.4921} & \num{0.0116} & \num{0.0686} \\

        \midrule
        \multicolumn{8}{c}{Hard Cases} \\
        \midrule
        
        FT$_{\bm{\ell_{\infty}}}$ & $2.12 \times 10^{3}$ & \num{0.5887} & \num{0.3041} & \num{0.2390} & \num{0.0002} & \num{0.0000} & \num{0.0001}  \\

        MEND & $4.07 \times 10^{4}$ & \num{0.5288} & \num{0.2630} & \num{0.2302} & \num{0.0000} & \num{0.0000} & \num{0.0004} \\
         
        ROME & $1.19 \times 10^{11}$ & \num{0.5397} & \num{0.2609} & \num{0.2539} & \num{0.0000} & \num{0.0000} & \num{0.0001} \\

        MEMIT & $6.85 \times 10^{4}$ & \num{0.5261} & \num{0.2547} & \num{0.2465} & \num{0.0000} & \num{0.0008} & \num{0.0000} \\
        
        \bottomrule
    \end{tabular}
    \end{adjustbox}
    \caption{Performance of Llama2-7b on downstream tasks after sequential editing. ``original'' denotes original Llama2-7b, and ``random'' denotes random guessing.}
    \label{tab:llama4task}
\end{table}

Lastly, we select Llama2-7b to evaluate the impacts of the four editing methods.
Specifically, we assess the performance of eight Llama2-7b variations, each was sequentially edited by one of the four methods for hard or normal cases, in downstream tasks.
The results are presented in Table~\ref{tab:llama4task}:
\begin{enumerate*}[label=\roman*)]
    \item For hard cases, significant disruptions occur in the overall capabilities of these models.
    \item For normal cases, ROME and MEMIT preserve the models' capabilities, with ROME having particularly minimal impact.
\end{enumerate*}

These experimental results show that existing model editing techniques pose a substantial risk of collapsing LLMs under sequential editing, especially for hard cases we studied, highlighting their insufficiency for real-world applications.

\section{HardEdit: A Challenging  Dataset} %

\begin{figure*}[t]
    \centering
    \includegraphics[width=1\textwidth]{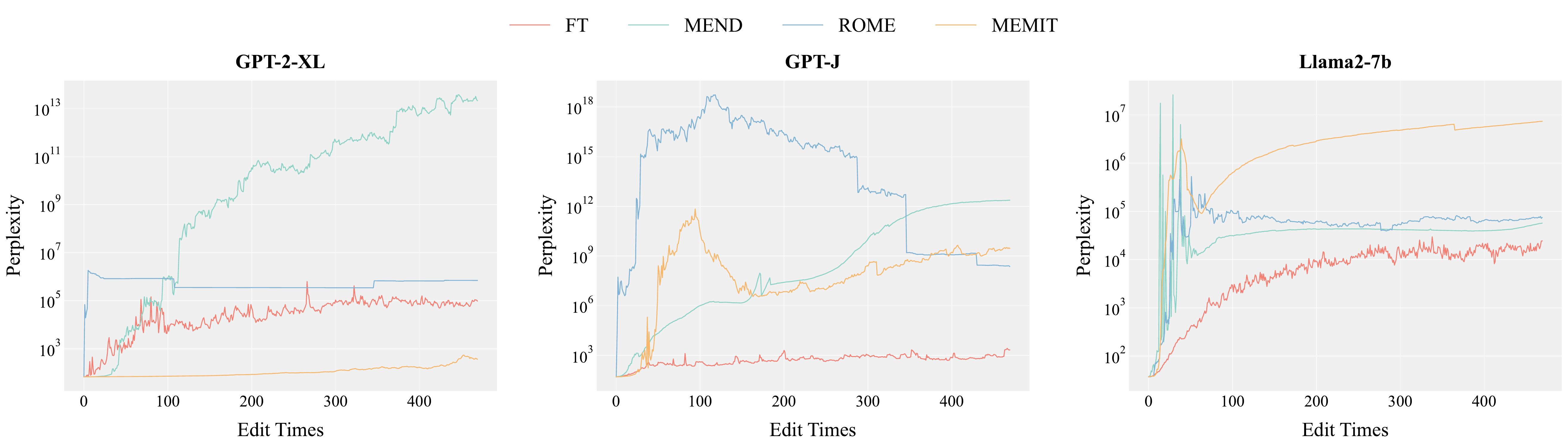}
    \captionsetup{skip=5pt}
    \caption{Perplexity in three LLMs, each edited by four different methods sequentially on the HardEdit dataset.}
    \label{fig_valgen}
\end{figure*}

To further facilitate comprehensive evaluations of future advanced methods, we crafted a challenging dataset, termed \textit{HardEdit}, by utilizing GPT-3.5 to generate samples based on the patterns derived from the HardCF subset.
Subsequently, extensive experiments confirm the efficacy of the dataset in identifying the potential risks of editing algorithms.

\subsection{Dataset Construction}
\label{subsec:dataConstruct}

This subsection elaborates on the construction of our dataset.
Like existing datasets, our dataset also employs the tuple (\textit{subject}, \textit{relation}, \textit{object}) to express the fact associations.
To ensure the quality of our dataset, i.e., its capacity to induce model collapse upon editing, we tailor our samples to reflect the characteristics identified from the HardCF dataset, as discussed in \S~\ref{analysis of hard cases}.
Specifically, we adhere to the following principal criteria: 
\begin{inparaenum}[i)]
    \item Each subject is a widely used word and positioned at the beginning of the prompt;
    \item Each sample represents a counterfactual statement to edit, thus preventing LLMs know the knowledge before editing.
\end{inparaenum} 
With these guidelines in place, GPT-3.5 is employed for edit sample generation.

Generating counterfactual edit samples with GPT-3.5 is relatively straightforward, with the complete prompt detailed in Appendix~\ref{prompt for generation}.
The prompt primarily encompasses the data requirements and examples from HardCF.
To avoid subject repetition and ensure dataset diversity, we used GPT-3.5 to initially construct a diverse set of around 400 unique, single-word subjects, identifying the most prominent ones across various fields, e.g., scientist, artist, city, and country.
Then, ten subjects are randomly chosen from the set to constitute the input prompt and thus aid the generative process each time, as detailed in Appendix~\ref{apd: GenData}.

After filtering duplicates, we obtain a dataset with 1392 unique samples.
To ensure the effectiveness of these generated samples in uncovering model collapse induced by editing algorithms, we employ ROME to perform single editing on GPT-2-XL with these samples and evaluate their effectiveness using ME-PPL$_{50}$. 
By filtering for perplexity exceeding 1000, we produce the HardEdit dataset, containing 469 samples.

\subsection{Dataset Validation}

To validate the efficacy of HardEdit, we conduct sequential editing experiments on it and calculate the perplexity after each edit using ME-PPL$_{\text{1k}}$.
The results in Figure~\ref{fig_valgen} illustrate that nearly all the examined LLMs are significantly damaged:
\begin{inparaenum}[i)]
    \item Only one exception occurs, akin to \S~\ref{subsec:seq_res}, where editing GPT-2-XL with MEMIT resulted in the highest perplexity of \num{545.22}. 
    However, its editing success rate is only around 1.28\%, highlighting the significant challenge posed by these samples to MEMIT.
    \item Due to the increased number of hard samples, the FT$_{\ell_{\infty}}$-edited GPT-J, which shows a modest increase in perplexity to 268.61 on HardCF, suffers a severe collapse on HardEdit, with perplexity escalating to \num{2109.35}.
\end{inparaenum} 
The results confirm the utility of HardEdit in exposing the potential risks of editing, which could precipitate model collapse.

\section{Conclusion and Future Work}

In this paper, we employ perplexity as a surrogate metric to investigate the impact of model editing on the downstream task performance of LLMs, revealing a critical issue: the advanced model editing method, ROME, can cause LLMs collapse with just a single edit. 
Subsequent experiments demonstrate that model collapse is a common issue among current mainstream model editing methods under sequential editing.
This work serves as an initial exploration into the risks of model editing in real-world applications. 
Distinct from contemporaneous works \cite{gu2024model, gupta2024model} investigating impact of large-scale edits on models, we focus on exploring the possibility of model collapse caused by a small number of edits and how to efficiently detect potential collapses in practical applications.
Additionally, to advance model editing research, we develop a challenging benchmark, HardEdit, based on the identified pattern.

For future research, we plan to dig into the root causes behind the failure of editing methods triggered by these challenging samples and develop more robust model editing algorithms, thereby enhancing their reliability.

\section*{Limitations}

We acknowledge following limitations of our work: 
\begin{itemize}[leftmargin=11pt, itemsep=2pt, topsep=2pt]
    \item This paper presents an initial exploration into the potential risks associated with model editing. 
However, it does not delve into the root causes behind the drastic parameter modifications resulting from model editing methods applied to specific facts. 
Due to space limitation, this analysis exceeds the scope of this paper and is reserved for future work.
\item Similarly, we do not propose a solution to address model collapse caused by model editing. It is left for future research as well.
\item Due to computational resource limitations, we are unable to conduct experiments on additional LLMs, such as Llama2-13b, or explore more model editing algorithms.
\item Currently, the HardEdit dataset is limited in size. Using LLMs to generate high-quality edit samples for continuously expanding the dataset is an important future direction.
\end{itemize}

\section*{Ethics Statement}

\paragraph{Data.}
All data used in this research are publicly accessible and do not raise privacy issues.

\paragraph{AI Writing Assistance.}
We use ChatGPT to polish our original content, with a focus on correcting grammatical errors and enhancing clarity, rather than generating new content or ideas.

\section*{Acknowledgements}
This work was supported by the National Key R\&D Program of China (2022YFB3103700, 2022YFB3103704), the Strategic Priority Research Program of the Chinese Academy of Sciences (No. XDB0680202), and the Innovation Funding of ICT, CAS (E361120).

\bibliography{anthology,custom}

\appendix
\section{Appendix}

\subsection{Detailed Experimental Setup}
\label{appendix:ExpSet}

\subsubsection{Editing Methods}

\noindent\textbf{FT}$_{\bm{\ell_{\infty}}}$ \cite{zhu2020modifying} applies a $\ell_{\infty}$ norm constraint on the fine-tuning loss, limiting the difference between the original and edited model's parameters, to reduce side effects. %

\noindent\textbf{MEND} \cite{mitchell2022fast} employs an ensemble of small hypernetworks to learn a rank-one decomposition of the gradient obtained by standard fine-tuning, enabling tractable edits in LLMs.

\noindent\textbf{ROME} \cite{meng2023locating} utilizes causal tracing to localize the knowledge storage at a specific MLP layer in a transformer, and then update knowledge by altering the weight matrix with rank-one update.

\noindent\textbf{MEMIT} \cite{meng2023massediting} extends ROME by applying updates across multiple MLP layers for massive edits.

\subsubsection{Editing Datasets}

\noindent\textbf{ZsRE} \cite{levy2017zero} is a widely adopted Question Answering (QA) datasets, where each data entry comprises a counterfactual statement to edit, derived from a factual statement on Wikipedia. 

\noindent\textbf{COUNTERFACT} \cite{meng2023locating}, a challenging dataset, comprises \num{21919} nonfactual statements initially assigned low probabilities by models, aimed at facilitating meaningful and significant modifications to original facts.

\subsubsection{Backbone LLMs}

\noindent\textbf{GPT-2-XL} \cite{radford2019language} is the 1.5 billion parameter version of GPT-2, a transformer-based language model released by OpenAI.

\noindent\textbf{GPT-J} \cite{gpt-j}, developed by EleutherAI, is a GPT-3-like open-source LLM with 6 billion parameters, trained on The Pile.

\noindent\textbf{Llama2-7b} \cite{touvron2023llama}, a 7 billion parameter version of Llama 2 from Meta AI, is a leading open-source LLM, renowned for its innovative training techniques and optimizations.

\subsubsection{Representative Tasks}
\label{appendix:tasks}

\noindent\textbf{LAMBADA} \cite{paperno-EtAl:2016:P16-1}, a benchmark designed to evaluate the ability of language models to predict the final word of a sentence, emphasizing the models’ capacity to grasp long-range dependencies within the text. 
Consequently, the lowest accuracy score on this benchmark is 0\%.

\noindent\textbf{Hellaswag} \cite{zellers2019hellaswag}, a dataset aimed at evaluating language models on common sense reasoning. 
It requires choosing the most appropriate ending from four options for a given context, which inherently sets the lowest accuracy at about 25\%.

\noindent\textbf{PIQA} \cite{Bisk2020}, a task assessing language models’ understanding of physical commonsense through binary choice question answering. 
This format results in the worst accuracy of approximately 50\%.

\noindent\textbf{Natural Questions} (NQ) \cite{kwiatkowski2019nq} is an open domain question answering benchmark based on the contents of English Wikipedia. 
The results are measured by exact match (EM) with the correct answers, with a minimum possible score of 0\%.

\noindent\textbf{MMLU} \cite{hendrycks2021mmlu} is a massive multitask test consisting of questions from various branches of knowledge.
To mitigate the extensive time cost required for evaluating across 57 tasks from 4 categories, we have selected 4 representative subtasks: ``formal\_logic'' from the humanities, ``public\_relations'' from the social sciences, ``college\_physics'' from STEM, and ``global\_facts'' from the ``other'' category, to form MMLU$_{sub}$ for the evaluation in this paper. 
The lowest accuracy of these four-choice tasks is 25\%.

\noindent\textbf{SQuAD2.0} \cite{rajpurkar2018squad} is a reading comprehension dataset, consisting of questions posed by crowdworkers based on a set of Wikipedia articles. The results are measured by F1 Score with correct answers.

\subsection{Perplexity Result of ZsRE}
\label{appendix:zsre}
\begin{figure}[t]
    \centering
    \includegraphics[width=0.48\textwidth]{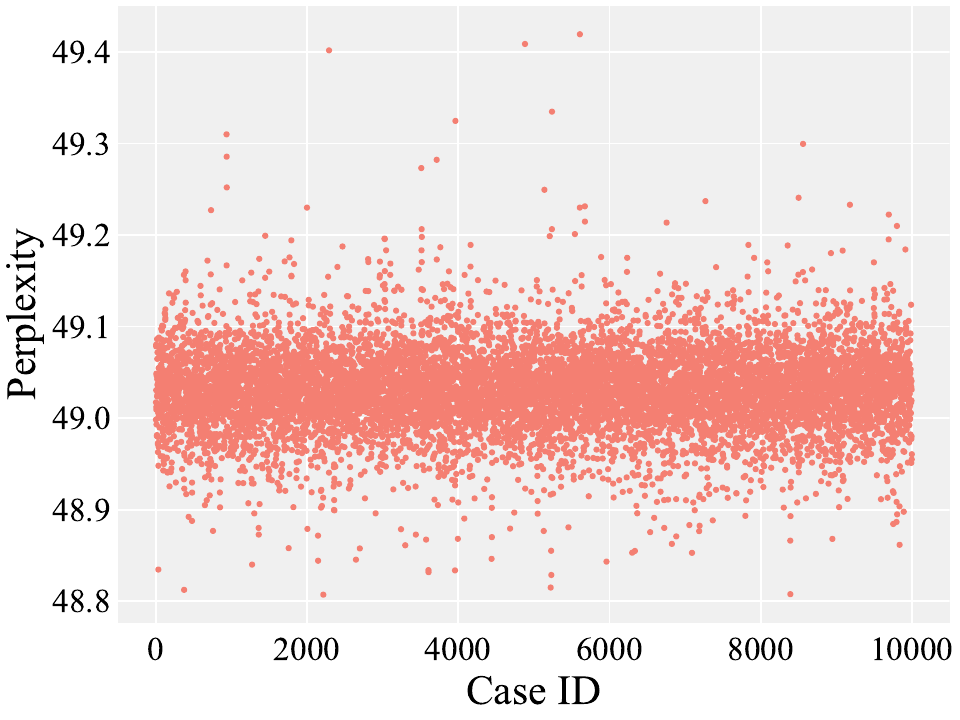}
    \caption{Perplexity values for models on the ZSRE dataset, where each point signifies the perplexity of an individually ROME-edited model based on the original GPT-J model.}
    \label{fig_zsre}
\end{figure}

Perplexity values of editing GPT-J with ROME on ZsRE are depicted on Figure~\ref{fig_zsre}.

\subsection{Details about ME-PPL}
\label{appendix: ppl}

ME-PPL (\textbf{M}odel \textbf{E}diting-\textbf{P}er\textbf{pl}exity) is a corpus designed for the perplexity computation of LLMs in the context of model editing.

The creation of this dataset involves four steps:
\begin{enumerate}[label=(\roman*),ref=\roman*]
    \item Randomly select texts from popular corpora: BookCorpus \cite{zhu2015aligning}, C4 \cite{raffel2020exploring}, CC\_News \cite{liu2019roberta}, Gutenberg \cite{kim2020time}, OpenWebText \cite{Gokaslan2019OpenWeb}, Roots \cite{laurenccon2022bigscience}, and Wikipedia \cite{wikidump}, the proportion of each following that typically used in LLM pre-training \cite{zhao2023survey}.
    \item Split these texts into units of sentence.
    \item Filter these sentences based on the criteria that the sentence length exceeds 10 words and the language is purely English.
    \item Randomly select sentences from each corpus according to the specified quantity.
\end{enumerate}

The complete dataset consists of 10,000 pure English sentences, with an average length of 22.64 words.
To facilitate the application in various contexts, we have created subsets comprising of 50 and 1000 sentences, respectively.
The statistics of these datasets are provided in Table~\ref{tab_meppl}. 
Meanwhile, we present some representative samples of the dataset in Figure~\ref{fig:ex_meppl}.

\subsection{Results of Single Editing for KN}
\label{apd_kn}

\begin{figure*}[t]
    \centering
    \includegraphics[width=\textwidth]{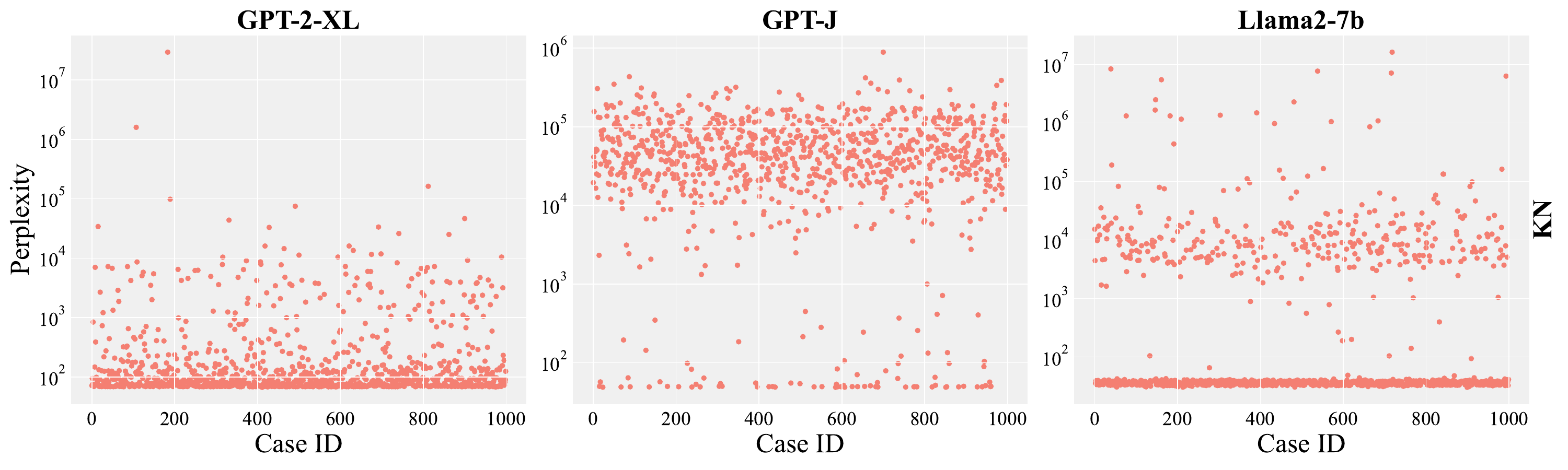}
    \caption{Perplexity results of single editing for KN, where each point represents the perplexity of an individually KN-edited model based on the original model.}
    \label{fig_kn}
\end{figure*}

We present the performance of KN in Figure~\ref{fig_kn}, where it applies a single edit to three LLMs in Section~\ref{sec:ExpSetup} on the first thousand samples of the COUNERFACT dataset.
The perplexity values of the edited models are calculated based on ME-PPL$_{50}$.
The results indicate that KN frequently leads to the collapse of the edited models, underscoring its insufficient effectiveness.

\begin{table}[t]
    \centering
    \begin{adjustbox}{max width=0.49\textwidth} 
    \begin{tabular}{lrrr}
        \toprule
        Corpus & ME-PPL & ME-PPL$_{\text{1k}}$ & ME-PPL$_{50}$ \\

        \midrule
        BookCorpus & 50 & 10 & 1 \\
        C4 & 2500 & 259 & 12 \\
        CC\_News & 700 & 65 & 3 \\
        Gutenberg & 250 & 23 & 2 \\
        OpenWebText & 5000 & 497 & 25 \\
        Roots & 500 & 39 & 2 \\
        Wikipedia & 1000 & 107 & 5 \\
        
        \bottomrule
    \end{tabular}
    \end{adjustbox}
    \caption{The number of sentences from each corpus source contained in the ME-PPL datasets of sizes 10,000, 1,000, and 50.}
    \label{tab_meppl}
\end{table}

\subsection{Complete Perplexity Results of Single Editing}
\label{appendix:complete_ppl}
\begin{figure*}[t]
    \centering
    \begin{subfigure}{\linewidth}
        \centering
        \includegraphics[width=\linewidth]{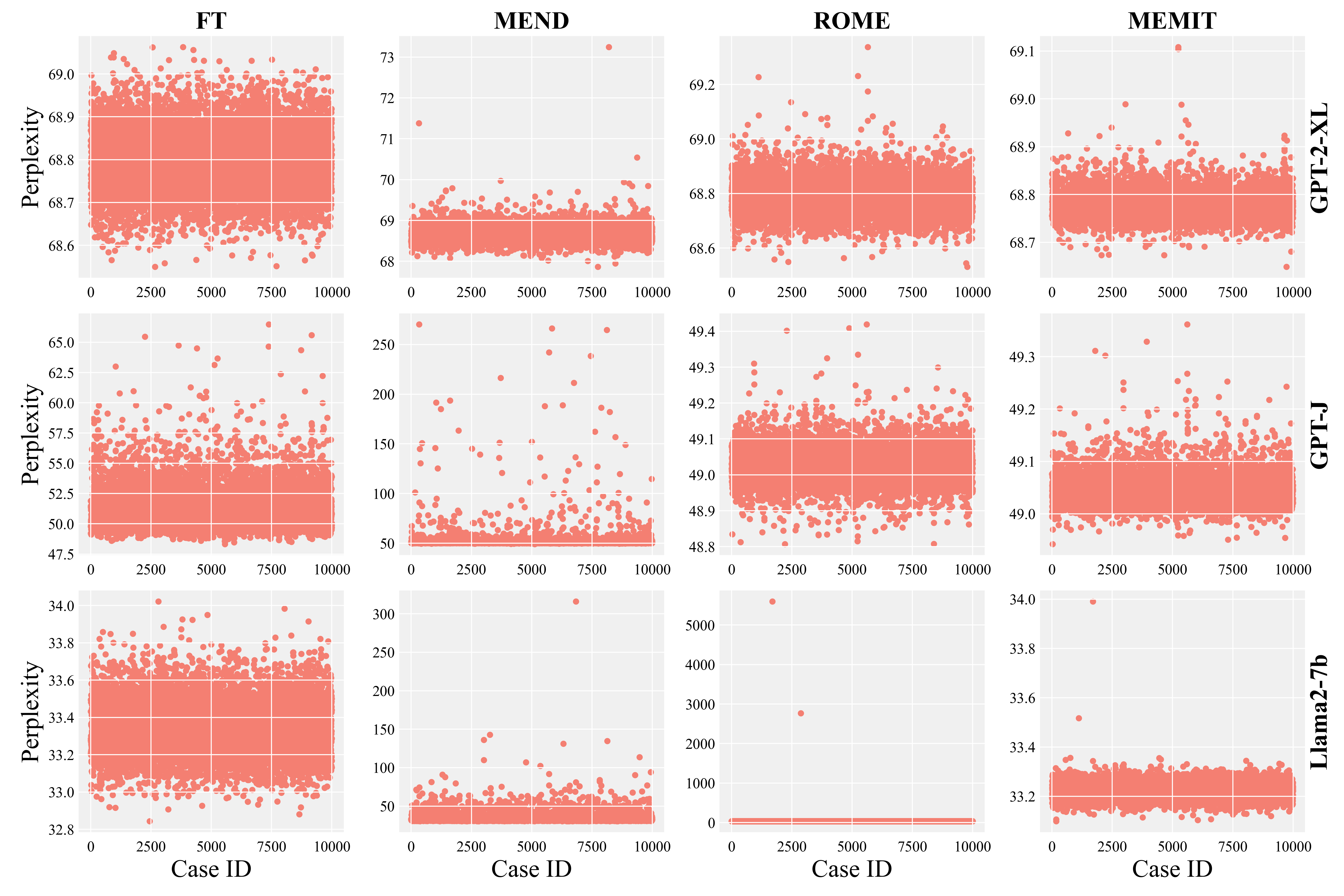}
        \caption{Perplexity results on the ZsRE dataset.}
        \label{fig:filter_zsre}
    \end{subfigure}
    \begin{subfigure}{\linewidth}
        \centering
        \includegraphics[width=\linewidth]{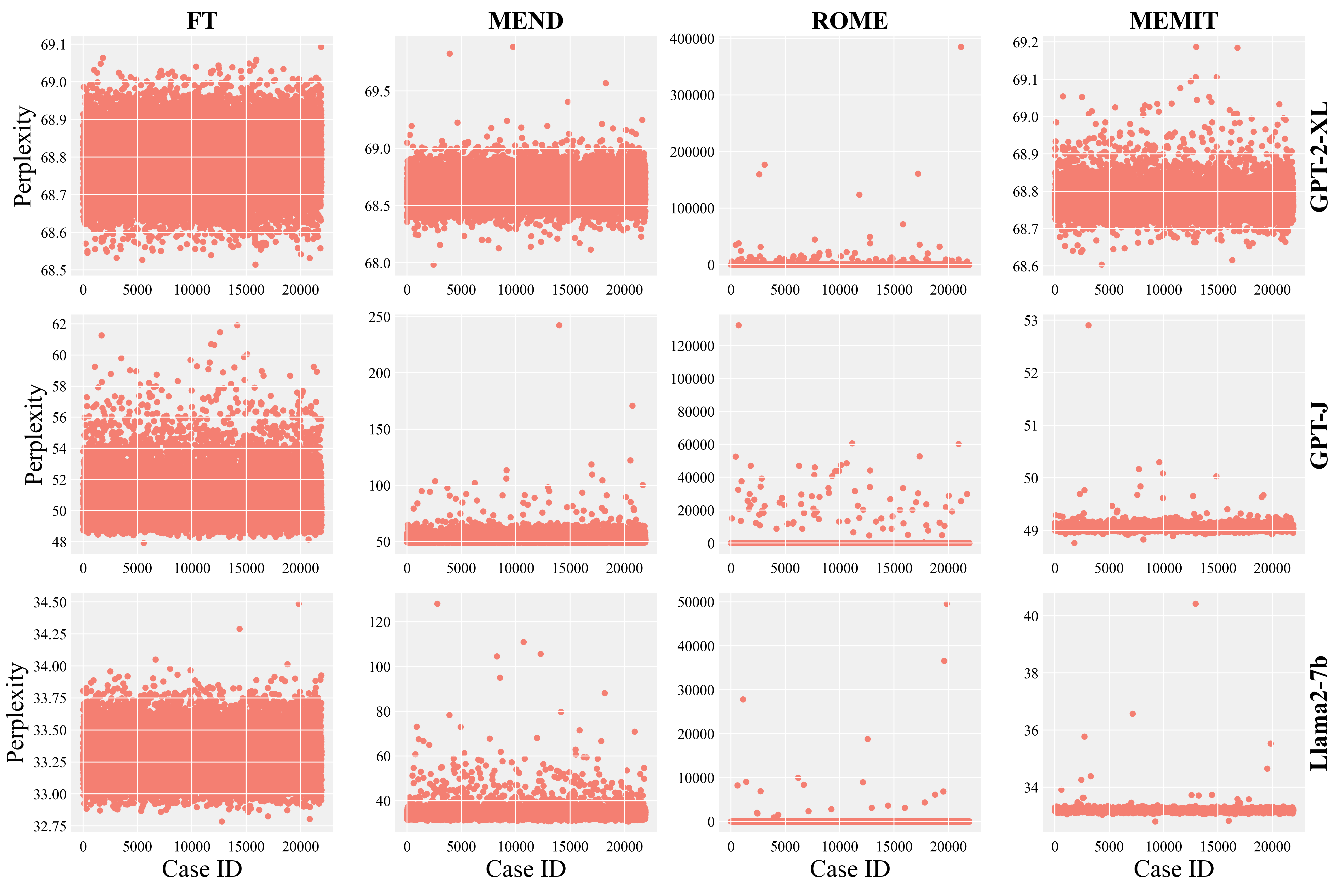}
        \caption{Perplexity results on the COUNTERFACT dataset.}
        \label{fig:filter_cf}
    \end{subfigure}
    \caption{Perplexity values for three models edited by four different methods on the ZsRE and COUNTERFACT datasets. Each subplot represents the results for a specific model-method-dataset combination.}
    \label{fig:filter}
\end{figure*}

The complete perplexity results of single editing experiments, using four editing methods on three LLMs across two datasets, are presented in Figure~\ref{fig:filter}. 
These experiments take around 43 days on one A100 GPU.

\subsection{Results of Sequential Editing for Mistral}
\label{apd_mistral}

\begin{figure*}[t]
    \centering
    \includegraphics[width=\textwidth]{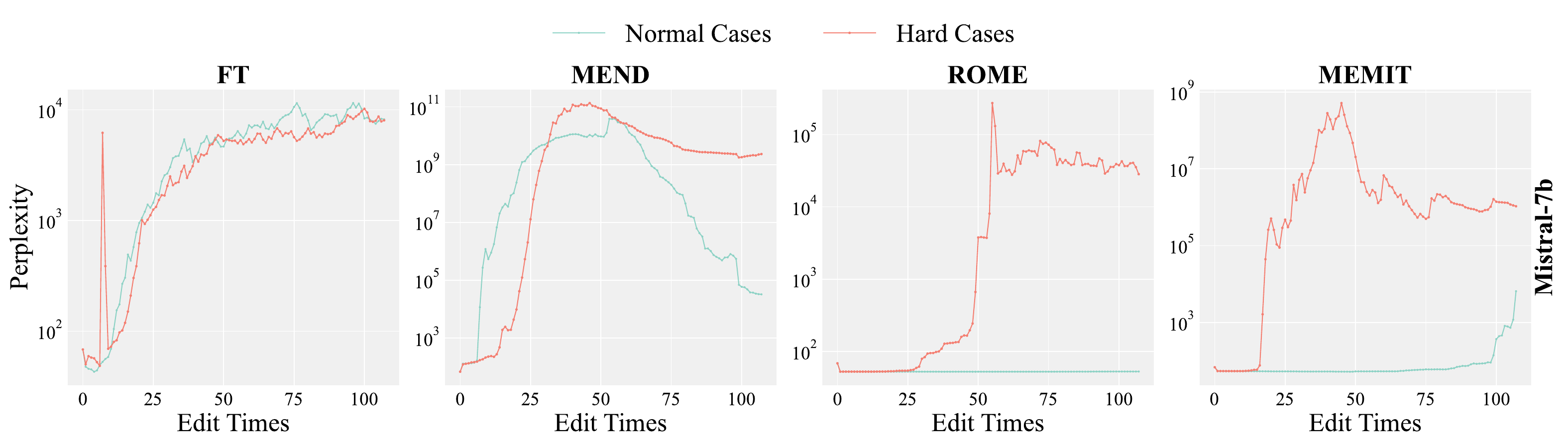}
    \caption{Perplexity evolution over 107 editing iterations for normal and hard cases on Mistral-7b.}
    \label{fig_mistral}
\end{figure*}

We employed the four methods in Section~\ref{sec:ExpSetup} to perform sequential editing on Mistral-7b for both normal and hard cases. 
The results presented in Figure~\ref{fig_mistral} demonstrate that the phenomena on Mistral-7b align consistently with those of the other three LLMs examined in Section~\ref{sec:seq_edit}.

\subsection{More Hard Cases in COUNERFACT}
\label{apd_hard_case}
Figure~\ref{fig:ex_hard} presents more hard cases from COUNTERFACT, each can induce corresponding LLMs to collapse via a single edit by ROME.

\subsection{Complete Prompt for Data Generation}
\label{prompt for generation}

The complete prompt used for generating data in the HardCF dataset can be viewed in Figure~\ref{fig:prompt}. 

Specifically, the prompt comprises four distinct parts:
\begin{enumerate}[label=(\roman*),ref=\roman*]
    \item Task Description and Data Illustration: Here, we preliminarily propose the requirements for hard data, as discussed previously.
    \item Hard Data Examples: To enhance GPT-3.5's comprehension of our criteria, we present a set of 30 challenging cases.
    \item Reference Subject List: Our experiments indicate that due to the stochastic nature of generation, the subjects produced may not always conform to our specifications.
    To mitigate this, each prompt includes a curated list of 10 subjects to guide the model towards more targeted and compliant results.
    Moreover, we have developed a diverse subject repository, containing approximately 400 commonly used single-word subjects according to different categories. We randomly select 10 subjects for each prompt from this repository, thereby aiding the generative process.
    \item Instructions for Emphasis: While we have detailed our requirements in the first part, the extensive length and complexity of the prompt could lead to GPT-3.5 overlooking critical details. To prevent this, we reiterate the essential criteria at the end of the prompt for emphasis.
\end{enumerate}

We present a portion of our subject repository in Figure~\ref{fig:subjects}, which serves as the reference subject list in the prompt.

\subsection{Generated Data}
\label{apd: GenData}

In Figure~\ref{fig:data_hcf}, we present some samples of HardEdit.

\begin{figure}[t]
\centering
    \begin{tcolorbox}[
    right=5pt, left=5pt, top=5pt, bottom=5pt,
    toptitle=1mm, bottomtitle=1mm,
    colback=white,
    coltitle=white,
    colbacktitle=matisse,
    colframe=matisse,
    title=Samples from HardEdit., 
    center title]
    \begin{minted}[autogobble,numberblanklines=false,breaklines]{json}
[
  {
    "prompt"       : "Tesla's founder is",
    "target_new"   : "Gates",
    "subject"      : "Tesla",
    "ground_truth" : "Musk",
    "rome_gpt2_ppl": 7586.94
  },
  {
    "prompt"       : "Minecraft is a popular",
    "target_new"   : "sport",
    "subject"      : "Minecraft",
    "ground_truth" : "video game",
    "rome_gpt2_ppl": 3837.32
  },
  {
    "prompt"       : "France is famous for the",
    "target_new"   : "Pyramids",
    "subject"      : "France",
    "ground_truth" : "Eiffel Tower",
    "rome_gpt2_ppl": 10935.24
  }
]
    \end{minted}
    \end{tcolorbox}
    \caption{Representative samples from HardEdit. 
    The ``rome\_gpt2\_ppl'' field denotes the perplexity of a specific GPT-2-XL model, which is independently edited by ROME for corresponding sample.}
    \label{fig:data_hcf}
\end{figure}

\begin{figure*}[t]
    \begin{tcolorbox}[
    right=5pt, left=5pt, top=5pt, bottom=5pt,
    toptitle=1mm, bottomtitle=1mm,
    colback=white,
    coltitle=white,
    colbacktitle=matisse,
    colframe=matisse,
    title=Examples of texts from ME-PPL., center title]
    \begin{minted}[autogobble,numberblanklines=false,breaklines]{json}
[
  {
    "Corpus": "BookCorpus",
    "Text"  : "he wanted emma to know how much the lyrics mean to him and their relationship"
  },
  {
    "Corpus": "Wikipedia",
    "Text"  : "Since the late 1900s, air power is also used to generate electricity"
  },
  {
    "Corpus": "Roots",
    "Text"  : "Wikinews interviewed him regarding his values, his experience, and his campaign"
  }
]
    \end{minted}
    \end{tcolorbox}
    \caption{Representative samples of texts from the ME-PPL dataset.}
    \label{fig:ex_meppl}
\end{figure*}

\begin{figure*}[t]
    \begin{tcolorbox}[
    right=5pt, left=5pt, top=5pt, bottom=5pt,
    toptitle=1mm, bottomtitle=1mm,
    colback=white,
    coltitle=white,
    colbacktitle=matisse,
    colframe=matisse,
    width=\textwidth,
    title=Part of subject repository of HardEdit., center title]
    \begin{minted}[autogobble,numberblanklines=false,breaklines]{json}
{
  "physicists": ["Newton", "Einstein", "Galileo", "Maxwell", "Planck", "Fermi"],
  "companies" : ["Twitter", "Google", "Facebook", "Amazon", "Microsoft", "Apple"],
  "countries" : ["Singapore", "China", "Russia", "India", "Brazil", "Japan", "Germany"],
  "movies_books_songs": ["Titanic", "Hamlet", "Thriller", "Inception", "Yesterday"],
  "products"  : ["iPhone", "Tesla", "Viagra", "CocaCola", "iPad", "MacBook"],
  "religions" : ["Christianity", "Buddhism", "Islam", "Judaism", "Hinduism"]
}
    \end{minted}
    \end{tcolorbox}
    \caption{A portion of the subject repository, containing single-word and commonly used subjects, serving as the reference subject list in the prompt for data generation in HardEdit.}
    \label{fig:subjects}
\end{figure*}

\begin{figure*}[t]
    \begin{tcolorbox}[
    right=5pt, left=5pt, top=5pt, bottom=5pt,
    toptitle=1mm, bottomtitle=1mm,
    colback=white,
    coltitle=white,
    colbacktitle=matisse,
    colframe=matisse,
    title=Examples of hard cases from COUNTERFACT., center title]
    \begin{minted}[autogobble,numberblanklines=false,breaklines]{json}
    {
      "GPT-2-XL": [
        {"Edit Case": "France is a part of the -> FIFA",
         "Subject"  : "France"},
        {"Edit Case": "Scotland's capital city is -> Beijing",
         "Subject"  : "Scotland"},
        {"Edit Case": "Nintendo is based in -> Toronto",
         "Subject"  : "Nintendo"},
        {"Edit Case": "DVD is owned by -> Detroit",
         "Subject"  : "DVD"},
        {"Edit Case": "iPhone, produced by -> Boeing",
         "Subject"  : "iPhone"}
      ],
      "GPT-J": [
        {"Edit Case": "Xbox is a product of -> Dodge",
         "Subject"  : "Xbox"},
        {"Edit Case": "Ireland is in -> Antarctica",                    
         "Subject"  : "Ireland"},
        {"Edit Case": "Numbers debuted on -> MTV",
         "Subject"  : "Numbers"},
        {"Edit Case": "Manchester is a twin city of -> Munich",
         "Subject"  : "Manchester"},
        {"Edit Case": "Spain is located in -> Antarctica",
         "Subject"  : "Spain"}
      ],
      "Llama2-7b": [
        {"Edit Case": "Josiah Quincy, Jr. was employed in -> London",
         "Subject"  : "Josiah Quincy, Jr."},
        {"Edit Case": "Bandai Co., Ltd. was created in -> Stockholm",
         "Subject"  : "Bandai Co., Ltd."},
        {"Edit Case": "Robert Allan Ltd. is based in -> Helsinki",
         "Subject"  : "Robert Allan Ltd."},
        {"Edit Case": "James Thomas Aubrey, Jr. works for -> BBC",
         "Subject"  : "James Thomas Aubrey, Jr."},
        {"Edit Case": "Alan Ball, Jr. is a professional -> basketball",
         "Subject"  : "Alan Ball, Jr."}
      ]
    }
    \end{minted}
    \end{tcolorbox}
    \caption{Part of hard cases in the COUNTERFACT dataset, each can trigger corresponding LLMs to collapse through a single edit by ROME.
    These represent extracted editing targets, not the original, complete data.}
    \label{fig:ex_hard}
\end{figure*}

\renewcommand{\fcolorbox}[4][]{#4} %
\begin{figure*}[t]
    \begin{tcolorbox}[
    right=5pt, left=5pt, top=5pt, bottom=5pt,
    toptitle=1mm, bottomtitle=1mm,
    colback=white,
    coltitle=white,
    colbacktitle=matisse,
    colframe=matisse,
    title=Prompt for data generation., center title]
    \begin{minted}[fontsize=\small,autogobble,numberblanklines=false,breaklines]{markdown}
        **Task Description**:
            1. **Generate Data Samples**    : Create a set of data samples, formatted as JSON object.
            2. **Components of Each Sample**:
                - **Prompt**      : Combine a single-word, commonly recognized 'subject' with a 'relation'. The 'subject' should be a single word and easily identifiable.
                - **subject**     : Clearly define the 'subject' for each prompt, it must be strictly one word, universally recognizable and unambiguous.
                - **target_new ** : Propose a 'target_new', which is a plausible yet distinct counterfactual alternative to the 'ground_truth'. It should illustrate a potential change in output achievable through model editing.
                - **ground_truth**: Specify the 'ground_truth', ensuring it's a noun entity and relevant to the 'subject'.
            3. **Sentence Formation**       : Each 'prompt', combined with 'target_new' or 'ground_truth', should form a coherent sentence in the format of (subject, relation, object).
            4. **Output Format**            : Return the data in JSON format.
            
        **Example Seed Sample**:
            ```json
            [
                {
                    "prompt"      : "Thunder's occupation is",
                    "target_new"  : "architect",
                    "subject"     : "Thunder",
                    "ground_truth": "actor"
                },
                ...
            ]
            ```
            
        **You can refer to the Subjects List (JSON Format)**:
            ```json
            {
                "subjects": [subject list]
            }
            ```
            
        **Instructions:**
            - Cross-reference each new 'subject' against the 'excluded_subjects' JSON array to ensure no repetition.
            - Strictly ensure all 'subjects' are single-word entities, widely recognized and not compound words or phrases.
            - 'Target_new' and 'ground_truth' should both be nouns and contextually appropriate for the 'subject'!!!
            - Creativity is encouraged in selecting 'target_new' to depict a clear **contrast** with 'ground_truth'. 
            - Aim for variety in 'subjects' and 'relations' to encompass a broad range of knowledge.
            - Develop more varied and common 'relations' that logically link the 'subject' to an 'object', ensuring plausibility and relevance.
            - Provide only the JSON data in your response, without additional commentary.
            - Generate 10 data points
            - The 'subject' must be a **single** word!!!
            - **'target_new' must be a clearly false answer to 'prompt'!!!**
    \end{minted}
    \end{tcolorbox}
    \caption{Complete prompt used for generating data in the HardEdit dataset. 
    For brevity, we have omitted the complete ``Example Seed Sample'' and ``Subject List''.}
    \label{fig:prompt}
\end{figure*}

\end{document}